\documentclass[a4paper,11pt]{article}
\pdfoutput=1
\usepackage[symbol]{footmisc}
\usepackage{hyperref}
\hypersetup{colorlinks, linkcolor={blue!40!black}, citecolor={blue!40!black}, urlcolor={blue!40!black}}

\providecommand{\keywords}[1]{\textbf{\textit{Index terms---}} #1}

\usepackage{xcolor}
\usepackage[]{dmhdr} 
\usepackage[square,sort,comma,numbers]{natbib}
\usepackage{enumitem}

\begin{document}
\title{\bcores: Robust Large-Scale Bayesian Data Summarization in the Presence of Outliers}
\author{\stepcounter{footnote}Dionysis Manousakas\thanks{Corresponding e-mail: \href{mailto:dm754@cam.ac.uk}{dm754@cam.ac.uk}} \and Cecilia Mascolo}
\date{%
		University of Cambridge
}
\maketitle
\begin{abstract}
	Modern machine learning applications should be able to address the intrinsic challenges arising over inference on massive real-world datasets, including scalability and robustness to outliers. Despite the multiple benefits of Bayesian methods (such as uncertainty-aware predictions, incorporation of experts knowledge, and hierarchical modeling), the quality of classic Bayesian inference depends critically on whether observations conform with the assumed data generating model, which is impossible to guarantee in practice. In this work, we propose a variational inference method that, in a principled way, can simultaneously scale to large datasets, and robustify the inferred posterior with respect to the existence of outliers in the observed data. Reformulating Bayes theorem via the \bdiv, we posit a robustified pseudo-Bayesian posterior as the target of inference. Moreover, relying on the recent formulations of Riemannian coresets for scalable Bayesian inference, we propose a sparse variational approximation of the robustified posterior and an efficient stochastic black-box algorithm to construct it. Overall our method allows releasing cleansed data summaries  that can be applied broadly in scenarios including structured data corruption. We illustrate the applicability of our approach in diverse simulated and real datasets, and various statistical models, including Gaussian mean inference, logistic and neural linear regression, demonstrating its superiority to existing Bayesian summarization methods in the presence of outliers. 
\end{abstract}
\section{Introduction}
\label{sec:introduction}

Machine learning systems perpetually collect growing datasets, such as product reviews, posting activity on social media, users feedback on services, or insurance claims. The rich information content of such datasets has opened up an exciting potential to tackle various practical problems. Hence, recent years have witnessed a surge of interest in scaling up inference in the large-data regime via stochastic and batch methods~\cite{angelino16, hoffman13, welling11}. Most of related approaches have treated datapoints indiscriminantly; nevertheless, it is well known that not all datapoints contribute equally valuable information for a given target task~\cite{ghorbani19}. 

Datasets collected in modern applications contain redundant input samples that reflect very similar statistical patterns, or multiple copies of identical observations. Often input aggregates subpopulations emanating from different distributions~\cite{zheng08, zhuang15}. Moreover, the presence of outliers is a ubiquitous challenge, attributed to multiple causes. In the first place, noise is inherent in most real-world data collection procedures, creating systematic outliers: crowdsourcing is prone to mislabeling~\cite{frenay13} and necessitates laborious data cleansing~\cite{lewis04, paschou10}, while measurements commonly capture sensing errors and system failures. Secondly, outliers can be generated intentionally from information contributing parties, who aim to compromise the functionality of the application through data poisoning attacks~\cite{barreno10, biggio12, li16, koh17, steinhardt17, ghorbani19}, realised for example via data generation from fake accounts. Outliers detection is  challenging, particularly in high dimensions~\cite{diakonikolas19, dickens20}. Proposed solutions \mbox{often are} model-specific, and include dedicated learning components which increase the time complexity of the application, involve extensive hyperparameter tuning, introduce data redundancies, or require model retraining~\cite{sheng08, whitehill09, raykar10, karger11, liu12, zhang16}. On the other hand, operating on a corrupted dataset is brittle, and can decisively degrade the predictive performance of downstream statistical tasks, deceptively underestimate model uncertainty and lead to incorrect decisions. 

In this work, we design an integrated approach for inference on massive scale observations that can jointly address scalability and data cleansing for complex Bayesian models, via robust data summarization. Our method inherits the full set of benefits of Bayesian inference and works for any model with tractable likelihood function. At the same time, it maintains a high degree of automation with no need for manual data inspection, no additional computational overhead due to robustification, and can tolerate a non-constant number of corruptions. Moreover, our work points to a more efficient practice in large-scale data acquisition, filtering away less valuable samples, and indicating the regions of the data space that are most beneficial for our inference task. 

Our solution can be regarded as an extension of Bayesian coreset methods that can encompass robustified inference. Bayesian coresets~\cite{huggins16, campbell19jmlr, campbell19neurips} have been recently proposed as a method that enables Bayesian learning at scale via substituting the complete dataset over inference with an informative sparse subset thereof. Robustified Bayesian inference methods~\cite{berger94} have sought solutions to mismatches between available observations and the assumed data generating model, %via generalizing the data likelihood function, 
via proposing heavy-tailed data likelihood functions~\cite{huber09, insua12} and localization~\cite{definetti61, wang18}, using robust statistical divergences~\cite{futami18, knoblauch18, miller19}, employing robust gradient estimates
over Langevin Monte Carlo methods~\cite{bhatia19}, or inferring datapoints-specific importance weights~\cite{wang17}. Here, we cast coreset construction in the framework of robustified inference, introducing \emph{\bcores{}}, a method that learns sparse variational approximations of the full data posterior under the \bdiv{}. In this way, we are able to yield summaries of large data that are distilled from outliers, or data subpopulations departing from our statistical model assumptions. Importantly, \bcores{} can act as a preprocessing step, and the learned data summaries can subsequently be given as input to any ordinary or robustified black-box inference algorithm.

The rest of this paper is organized as follows. In \cref{sec:preliminaries,sec:method} we introduce necessary concepts from Bayesian inference, and present our proposed method. In \cref{sec:evaluation} we expose experimental results on simulated and real-world benchmark datasets: we consider diverse statistical models and scenarios of extensive data contamination, and demonstrate that, in contrast to existing summarization algorithms, our method is able to maintain reliable predictive performance in the presence of structured and unstructured outliers. Finally, in \cref{sec:conclusion} we provide conclusions and discuss future works.

\section{Preliminaries}
\label{sec:preliminaries}

In this section, we introduce the required concepts from Bayesian inference, present robustness limitations of standard posterior on big data, and outline existing generalizations of the posterior that aim to robustify inference with respect to data mismatch.

\subsection{Standard Bayesian inference and lack of robustness in the large-data regime}
In the context of Bayesian inference, we are interested in updating our beliefs about a vector of random variables $\theta \in \Theta$, initially expressed through a prior distribution $\pi_0(\theta)$, after observing a set of datapoints $ x:=(x_n)_{n=1}^{N} \in \mcX^N$.  Posterior on $\theta$ can be computed via the application of Bayes rule 
\[
\pi(\theta|x) = \frac{1}{Z'}\pi(x|\theta)\pi_0(\theta),
\label{eq:bayes-rule}
\] 
where $Z'$ is a (typically intractable) normalization constant, and $\pi(x|\theta)$ is the likelihood of our observations according to an assumed statistical model.
When datapoints are conditionally independent given $\theta$---which is the primary focus of this work---likelihood gets factorized as $\pi(x|\theta) = \Pi_{n=1}^{N}\pi(x_n|\theta)$. An equivalent formulation of  the
Bayesian posterior as a solution to an optimization problem was proposed by Zellner~\citep{zellner88}, which is written as  
\[
\pi(\theta|x) = \frac{1}{Z'} \exp\left(-\xent{\hpi(x)}{\pi(x|\theta)}\right)\pi_0(\theta).
\label{eq:zellner-rule}
\]
In the above, $\hpi(x)$ is the empirical distribution of the observed datapoints. The exponent $\xent{\hpi(x)}{\pi(x|\theta)}:=-\sum_{n=1}^{N}\log\pi(x_n|\theta)$
corresponds (up to a constant) to the \emph{cross-entropy}, which is  equal to the empirical average of negative log-likelihoods of the datapoints, and quantifies the expected loss incurred by our estimates for the model parameters $\theta$ over the available observations, under the \emph{Kullback-Leibler~(KL) divergence}.

When $N$ is large, the Bayesian posterior is strongly affected by perturbations in the observed data space. To develop an intuition on this, assuming that the true and observed data distributions have densities $\pi_\theta$ and $\pi_\textsubscript{obs}$ respectively, we can rewrite an approximation of~\cref{eq:zellner-rule} via the KL divergence ($\plainkl$) as~\cite{miller19}
\[
\pi(\theta|x) 
& \propto  \exp\left(\sum_{n=1}^{N}\log\pi(x_n|\theta)\right) \pi_0(\theta)
\doteq	 \exp\left(N \int \pi_\textsubscript{obs} \log \pi_\theta\right) \pi_0(\theta) \\
 &:= \exp\left(-N \kl{\pi_\textsubscript{obs}}{\pi_\theta}\right) \pi_0(\theta),
 \label{eq:missmatch_with_N}
\]
where $\doteq$ denotes  agreement to first order in exponent.\footnote{i.e. $a_n \doteq b_n$ iff $(1/n)\log(a_n/b_n) \rightarrow 0$}
Hence, due to the large $N$ in the exponent, small changes to $\pi_\textsubscript{obs}$ will have a large impact on the posterior.

\subsection{Robustified posteriors}

Robust inference methods aim to adapt~\cref{eq:bayes-rule} to formulations that can address the case of observations departing from model assumptions, as often happening in practice, e.g. due to misspecified shapes of data distributions and number of components, or due to the presence of outliers. In such formulations~\citep{dawid16, jewson18, fujisawa08, eguchi01}, Bayesian updates rely on utilising robust divergences instead of the KL divergence, to express the losses over the data. 

A popular choice~\citep{futami18, knoblauch18} for enhancing robustness of inference is replacing the log-likelihood terms arising in~\cref{eq:zellner-rule} with the \emph{\bdiv~}(or \emph{density power divergence})~\citep{basu98, cichocki10}, which yields the following posterior for $\theta$~\citep{ghosh16,knoblauch18}
\[
\pi_\beta(\theta|x) \propto \exp\left(-\db{\hpi(x)}{\pi(x|\theta)}\right)\pi_0(\theta),
\label{eq:b-posterior}
\] 
where 
\[
\db{\hpi(x)}{\pi(x|\theta)} := 
 -\sum_{n=1}^{N}  \underbrace{\left(\frac{\beta+1}{\beta}\pi(x_n|\theta)^{\beta} + \int_{\mcX} \pi(\chi|\theta)^{1+\beta}d\chi\right)}_{:=f_n(\theta)},
\label{eq:b-loss}
\]
with $\beta>0$.
We refer to quantities defined in~\cref{eq:b-posterior,eq:b-loss} as the \emph{\bpost{}} and \emph{\blik{}} respectively. Noticeably, the individual terms $f_n(\theta)$  of the \blik{} 
allow attributing \emph{different strength of influence to each of the datapoints}, depending on their accordance with the model assumptions. As densities get raised to a suitable power $\beta$, outlying observations are exponentially downweighted. When $\beta \rightarrow 0$,~\cref{eq:zellner-rule} is recovered and all datapoints are treated equally.

In the presentation above we focused on modeling observations $(x_n)_{n=1}^{N}$~(unsupervised learning). In the case of supervised learning on data pairs ${(x_n,y_n)}_{n=1}^{N} \in (\mcX \times \mcY)^N$, the respective expression for individual terms of \blik{}\footnote{In this context for simplicity we use notation $f_n(\cdot)$  to denote $f(y_n|x_n, \cdot)$.} is~\cite{basu98}
\[
f_n(\theta):= -\frac{\beta+1}{\beta}\pi(y_n|x_n,\theta)^{\beta} +  \int_{\mcY} \pi(\psi|x_n,\theta)^{1+\beta}d\psi.
\label{eq:sl-lik-terms}
\] 
\section{Method}
\label{sec:method}

In this section we discuss \bcores, our unified solution to the robustness and scalability challenges of large-scale Bayesian inference. \cref{subsec:sparse-b-posterior} introduces the main quantity of interest in our inference method, and shows how it addresses the exposed issues. \cref{subsec:bb-construct} presents an iterative algorithm that allows efficient approximate computations of our posterior.

\subsection{Sparse $\bpost{}$}
\label{subsec:sparse-b-posterior}

Scaling up the computation of~\cref{eq:b-posterior} in the regime of massive datasets for non-conjugate models is challenging:
%~Closed form computation of~\bdiv{} is not possible for a large number of statistical models. Moreover, 
similarly to  ~\cref{eq:bayes-rule}, applying Markov chain Monte Carlo~(MCMC) methods to sample from the~\bpost{}, implies a computational cost scaling at order $\Theta(N)$. 

Bayesian coresets~\cite{huggins16,campbell19jmlr} have been recently proposed as a method to circumvent the computational cost for the purposes of approximate inference via summarizing the original dataset  $(x_n)_{n=1}^{N}$ with a small learnable subset of weighted datapoints $(x_m, w_m)_{m=1}^{M}$, where  $(w_m)_{m=1}^{M} \in \reals^{M}_{+},\; M \ll N$. 
Substituting~\cref{eq:b-loss} in ~\cref{eq:b-posterior}, allows us to explicitly introduce a weights vector $ w \in \reals_{\geq 0}^{N}$ in the posterior, and rewrite the latter in the general form
\[
\pi_{\beta,w}(\theta|x) 
= \frac{1}{Z(\beta, w)}  \exp\left(\sum_{n=1}^{N}w_nf_n(\theta)\right)\pi_0(\theta).
\label{eq:bcore-posterior}
\]
In the case of the \bpost{} on the full dataset~\cref{eq:b-posterior}, we have $w=\vecone \in\reals^{N}$; for coreset posteriors this vector acts as a learnable parameter and attains a non-trivial sparse value, with non-zero entries corresponding to the elements of the full dataset that are selected over the summarization.

Although Bayesian coresets can dramatically reduce inference time, they inherit the susceptibility of Bayesian posterior to data mismatch in the large data regime: even though the number of points used in inference gets reduced, these points are now weighted, hence the remark of~\cref{eq:missmatch_with_N} can carry over in coresets posterior. 

The recent formulation of Riemannian coresets~\citep{campbell19neurips} has framed the problem of coreset construction as Variational Inference~(VI) in a sparse exponential family. Our method provides a natural extension of this framework to robust divergences. Here we aim to approximate data posterior via a \emph{sparse \bpost}, which can be expressed as follows
\[
w^{*} = \arg \min_{w\in\reals^{N}} \kl{\pi_{\beta,w}}{\pi_{\beta}} 
\quad
\text{s.t.}
\quad
w \geq 0,\; ||w||_0 \leq M,
\label{eq:coreset-vi}
\]
In the following we denote expectations and covariances under $\theta \sim \pi_{\beta,w}(\theta|x)$ as $\EE_{\beta,w}$ and $\cov_{\beta,w}$ respectively. Then the KL divergence is written as
\[
\kl{\pi_{\beta,w}}{\pi_\beta}:=\EE_{\beta, w} \left[\log\frac{\pi_{\beta,w}}{\pi_\beta}\right].
\label{eq:kld}
\]
In our formulation it is easy to observe that posteriors of~\cref{eq:bcore-posterior} form a set of \emph{exponential family distributions}~\cite{wainwright08}, with natural parameters $w \in \reals_{\geq 0}^N$, sufficient statistics $(f_n(\theta))_{n=1}^{N}$, and log-partition function $\log Z(\beta, w)$. Following~\citep{campbell19neurips}, the objective can be expanded as 
\[
\kl{\pi_{\beta,w}}{\pi_\beta} =& \log Z(\beta) - \log Z(\beta, w) \\
                                    &- \sum_{n=1}^{N}\EE_{\beta,w}\left[f_n(\theta)- w_n f_n(\theta)\right],
\label{eq:sparsevi-obj}
\]
and minimized via gradient descent on %$\beta$ and 
$w$.  %As detailed in~\cref{sec:gradient-derivations}, 
The gradient of the objective of~\cref{eq:sparsevi-obj} can be derived in closed form, as  
\[
\nabla_{w}\kl{\pi_{\beta,w}}{\pi_\beta} 
												   & = -\cov_{\beta,w}\left[f,(1 -w)^Tf\right], 
\label{eq:dkl-grad}
\]
where $f:=\left[f_1(\theta) \ldots f_N(\theta)\right]^T$.%, $f':=\left[f'_1(\theta) \ldots f'_N(\theta)\right]^T$.

\subsection{Black-box stochastic scheme for incremental coreset construction}
\label{subsec:bb-construct}

\begin{algorithm*}[!t]
	\caption{Incremental construction of sparse \bpost{}}
	\label{alg:bcores}
	\begin{algorithmic}[1]
		\Procedure{\bcore}{$f,  \pi_0, x, M, B, S, T, (\gamma_t)_{t=1}^{\infty}$, $\beta$}
		\State $w \assign \mathbf{0} \in \reals^{M}$,\;\; $g \assign \mathbf{0} \in \reals^{S \times M}$, \;\;$g' \assign \mathbf{0} \in \reals^{S \times B}$, \;\;$\mcI \assign \emptyset$
		\For{$m =1, \ldots, M $}
		\LineCommentIndent{Take S samples from current coreset posterior}
		\State $(\theta)_{s=1}^{S}  \distiid \pi_{\beta,w} \propto \exp \left(w^Tf\right) \pi_0(\theta)$
		\LineComment{Obtain a minibatch of B datapoints from the full dataset}
		\State $\mcB\dist\distUnifSubset\left([N], B\right)$
		\LineComment{Compute the \blik{} vectors over the coreset and minibatch datapoints for each sample}
		\State $g_{s} \gets \left( f(x_m, \theta_s, \beta ) - \frac{1}{S}\sum_{r=1}^S f(x_m, \theta_{r}, \beta) \right)_{m \in \mcI}\in \reals^M$ 
		\State  $g'_{s} \gets \left( f(x_b, \theta_s, \beta) - \frac{1}{S}\sum_{r=1}^S f(x_b, \theta_{r}, \beta) \right)_{b\in\mcB} \in \reals^B$
		\LineComment{Get empirical estimates of correlation over the coreset and minibatch datapoints}
		\State ${\hcorr \assign \diag \left[ \frac{1}{S} \sum_{s=1}^{S} g_{s}
		{g_{s}}^T\right]^{-\frac{1}{2}} \left(\frac{1}{S} \sum_{s=1}^{S} g_{s} \left(\frac{N}{B}1^T g'_{s} - w^T g_{s}\right)\right) \in \reals^{M}}$
		\label{lst:line:core-corr}
		\State ${\hcorr' \assign \diag \left[ \frac{1}{S} \sum_{s=1}^{S} g'_{s}
		{g'_{s}}^T\right]^{-\frac{1}{2}} \left(\frac{1}{S} \sum_{s=1}^{S} g'_{s} \left(\frac{N}{B}1^T g'_{s} - w^T g_{s}\right)\right)  \in \reals^{B}}$
		\label{lst:line:batch-corr}
		\LineComment{Add next datapoint via correlation maximization}
		\State ${n^{\star} \assign \arg \underset{n \in [m] \cup [B]}{\max} \left( \left|\hcorr \right| \cdot \vecone[n \in \mcI] + \hcorr' \cdot \vecone [n \notin \mcI]\right), \; \mcI \assign \mcI \cup \{n^{\star}\}}$
		\LineComment{Optimize weights vector via projected gradient descent}
		\For{$ t = 1, \ldots, T$} 		
		\State $(\theta)_{s=1}^{S}  \distiid \pi_{\beta,w}(\theta) \propto \exp\left(w^T f\right)\pi_0(\theta)$ 
		\State $\mcB\dist\distUnifSubset\left([N], B\right)$
		\For{$s = 1, \dots, S$} 
		\LineComment{Compute gradient terms discretizations over the coreset and minibatch datapoints for each sample}
		\State $g_{s} \gets \left( f(x_m, \theta_s, \beta) - \frac{1}{S}\sum_{r=1}^S f(x_m, \theta_{r}, \beta) \right)_{m\in\mcI} \in \reals^M$  
		\State $g'_{s} \gets \left( f(x_b, \theta_s, \beta) - \frac{1}{S}\sum_{r=1}^S f(x_b, \theta_{r}, \beta) \right)_{b\in\mcB} \in \reals^B$  	
		\EndFor
		\LineComment{Compute MC gradients for variational parameters}
		\State $\hat\nabla_w \gets -\frac{1}{S}\sum_{s=1}^S g_s\left( \frac{N}{B} 1^Tg'_s- w^Tg_s\right)$
		%, $\hat\nabla_\beta \gets -w^T\frac{1}{S}\sum_{s=1}^S k_s\left( \frac{N}{B} 1^Tg'_s- w^Tg_s\right)$
		\label{lst:line:mc-grad}
		\LineComment{Take a projected stochastic gradient step}
		\State $w \gets \max(w - \gamma_t\hat\nabla_w, 0)$
		\EndFor
		\EndFor
		\State\Return $w$%, $\beta$
		\EndProcedure		 
	\end{algorithmic}
\end{algorithm*}

To scale up coreset construction on massive datasets we use stochastic gradient descent on minibatches $\mcB \sim \distUnifSubset([N], B)$, with $B \ll N$.
The covariance of~\cref{eq:dkl-grad} required for exact gradient computation of the variational objective is generally not available in analytical form. Hence, for our black-box coreset construction we approximate this quantity via Monte Carlo estimates, using samples of the unknown parameters from the coreset posterior. These samples can be efficiently obtained with complexity $O(M)$ (not scaling with dataset size $N$) due to the sparsity of the coreset posterior over the procedure. The proposed black-box construction makes no assumptions on the statistical model other than having tractable \blik{}s. We employ a two-step incremental scheme, with complexity of order $O\left(M(M+B)ST\right)$, where $S$ is the number of samples from the coreset posterior, and $T$ is the total number of iterations over coreset points weights optimization. The full incremental construction is outlined in \cref{alg:bcores}.

\subsubsection{Next datapoint selection}
We first select the next datapoint to include in our coreset summary, via a greedy selection criterion. Although maximizing decrease in KL locally via~\cref{eq:dkl-grad}, seems to be the natural greedy choice here, using the information-geometric argument presented in~\cite{campbell19neurips}, we use instead the following correlation maximization criterion:
\[
x_{m} = \arg \underset{x_m \in \mcI \cup \mcB }{\max}
\begin{cases}
	\left|\corr_{\beta,w}\left[f_m, \frac{N}{B}1^Tf - w^T f\right]\right| & w_m>0 \\
	\corr_{\beta,w}\left[f_m, \frac{N}{B}1^Tf - w^T f\right] & w_m=0,
\end{cases}
\label{eq:greedy-select}
\]
where we denoted by $\mcI$ the set of coreset points.
The correlations for coreset and minibatch datapoints are empirically approximated  as in lines~\ref{lst:line:core-corr} and~\ref{lst:line:batch-corr} of~\cref{alg:bcores} respectively.

\subsubsection{Coreset points reweighting}
After adding a new datapoint we update the coreset weight vector $w\in\reals_{\geq 0}$ via $T$ steps of projected stochastic gradient descent, using the Monte Carlo estimate of~\cref{eq:dkl-grad} per line~\ref{lst:line:mc-grad} of~\cref{alg:bcores}.
\vspace{.2cm}
\par 
\textbf{Summarization of observations groups and batches.}
Apart from working at the individual datapoints level, our scheme also enables summarizing batches and groups of observations. Acquiring efficiently informative batches of datapoints can replace random minibatch selection commonly used in stochastic optimization for large-scale model training. This extension can also be quite useful in situations where datapoints are partitioned in clusters, e.g. according to demographic information. For example, when gender and age features are available in datasets capturing users movies habits, collected datapoints can be binned accordingly, and our group summarization technique will allow extracting informative combinations of demographic groups that can jointly summarize the entire population's information. The robustness properties of \bcores{} in such applications can aid removing group bias, and rejecting groups with large fractions of outliers. \cref{alg:bcores} is again directly applicable, where $g_s$ vectors are now summed over the corresponding datapoints of each batch or group.

\section{Experiments \& Applications}
\label{sec:evaluation}
\definecolor{darkblue}{rgb}{0.0, 0.0, 0.55}
\definecolor{oxfordblue}{rgb}{0.0, 0.13, 0.28}
\newcommand{\MYhref}[3][oxfordblue]{\href{#2}{\color{#1}{#3}}}%

We examine the inferential results achieved by our method under 3 statistical models, in scenarios capturing different types of data mismatch with reality. The data contamination models used in following experiments are reminiscent of \emph{Huber's $\eps$-contamination model}~\citep{huber92}, which postulates that observed data are generated from a mixture of distributions of the form $(1-\eps)\cdot G+ \eps\cdot Q$, where $\eps \in (0,1)$,  $G$ is a distribution of inliers captured by the assumed statistical model, and $Q$ is an arbitrary distribution of outliers. This model has found use in several recent studies on robust statistical estimators suitable for underlying distributions with minimal assumptions~\citep{wei17, chen18}.

 \bcores{} is compared against a uniformly random sampling baseline, and stochastic batch implementations of two existing Riemannian coreset methods: 
 \benum[label={(\roman*)}]
 \item \sparsevi~\cite{campbell19neurips}, which builds up a coreset according to an incremental scheme similar to ours, considering the standard likelihood function terms evaluated on the dataset points, and 
 \item \psvi~\cite{psvi}, which runs a batch optimization on a set of pseudopoints, and uses standard likelihood evaluations to jointly learn the pseudopoints weights and locations so that the extracted summary resembles the statistics of the full dataset. 
\eenum

We default the number of iterations in the optimization loop over gradient-based coreset constructions to $ T = 500$, using a learning rate $ \gamma_t \propto t^{-1}$ and $S=100$ random projections per gradient computation. For consistency with the compared baselines, we evaluate inference results obtained by \bcores{} using the classical Bayesian posterior from~\cref{eq:bayes-rule} conditioned on the corresponding robustified data summary. Additional details on used benchmark datasets are presented in~\cref{sec:data-details}. Code is available at 	\MYhref{https://github.com/dionman/beta-cores}{https://github.com/dionman/beta-cores}.

\subsection{Simulated Gaussian Mean Inference under Stuctured Data Contamination}
\label{subsec:gauss-expt}

In this experiment we study how \bcores{} behaves in the setting of mean inference on synthetic $d$-dimensional data, sampled \iid from a normal distribution with known covariance,
\[
\theta \sim \distNorm\left(\mu_0, \Sigma_0\right),
\qquad 
\;\;\;\;\;
x_n \distiid \distNorm(\theta, \Sigma),
\qquad
n = 1, \ldots, N.
\label{eq:mvn-expt}
\]
In the presented results, we use priors $\mu_0=\mathbf{0}$ and $\Sigma_0=I$,  dimensionality $d=20$ and dataset size $N=5,000$.
 
We consider the case of structured data corruption existing in the observations, simulated as follows: Observed datapoints are typically sampled from a Gaussian $ \distNorm(\mathbf{1}, I)$. At a percentage $F\%$,  data collection fails; in this case, datapoints are collected from a shifted Gaussian $ \distNorm(\mathbf{10}, I)$. Consequently, the observed dataset forms a Gaussian mixture with two components; however, our statistical model assumes only a single Gaussian.

\begin{figure}[t!]
	\centering 
	\begin{subfigure}[b]{0.9\textwidth} 
		\includegraphics[width=1.\textwidth]{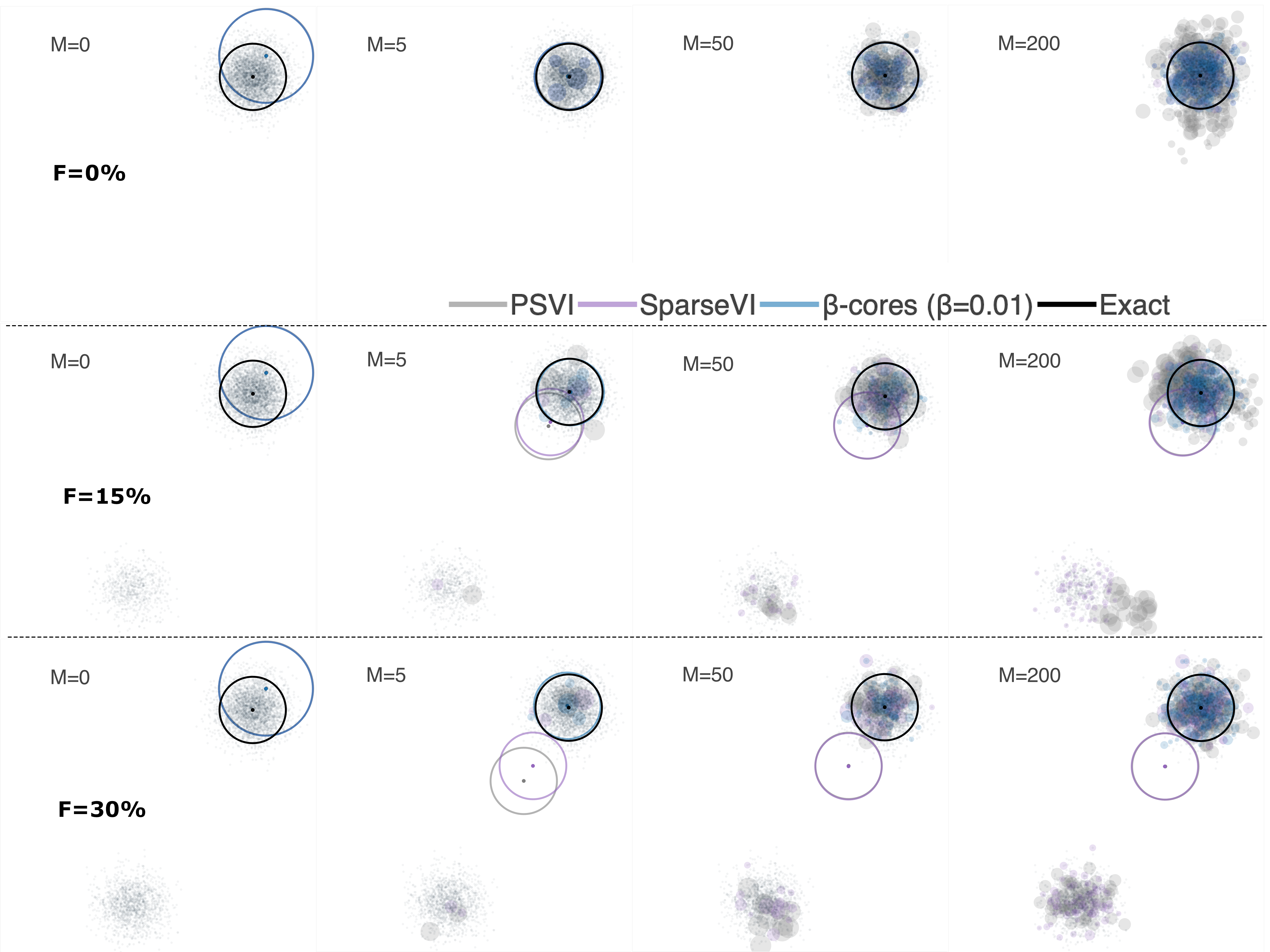}
		\caption{\label{fig:gaussian_coreset_points}}
	\end{subfigure}
	\hfill\qquad
	\begin{subfigure}[b]{0.9\textwidth} 
		\centering
		\includegraphics[width=.3\textwidth]{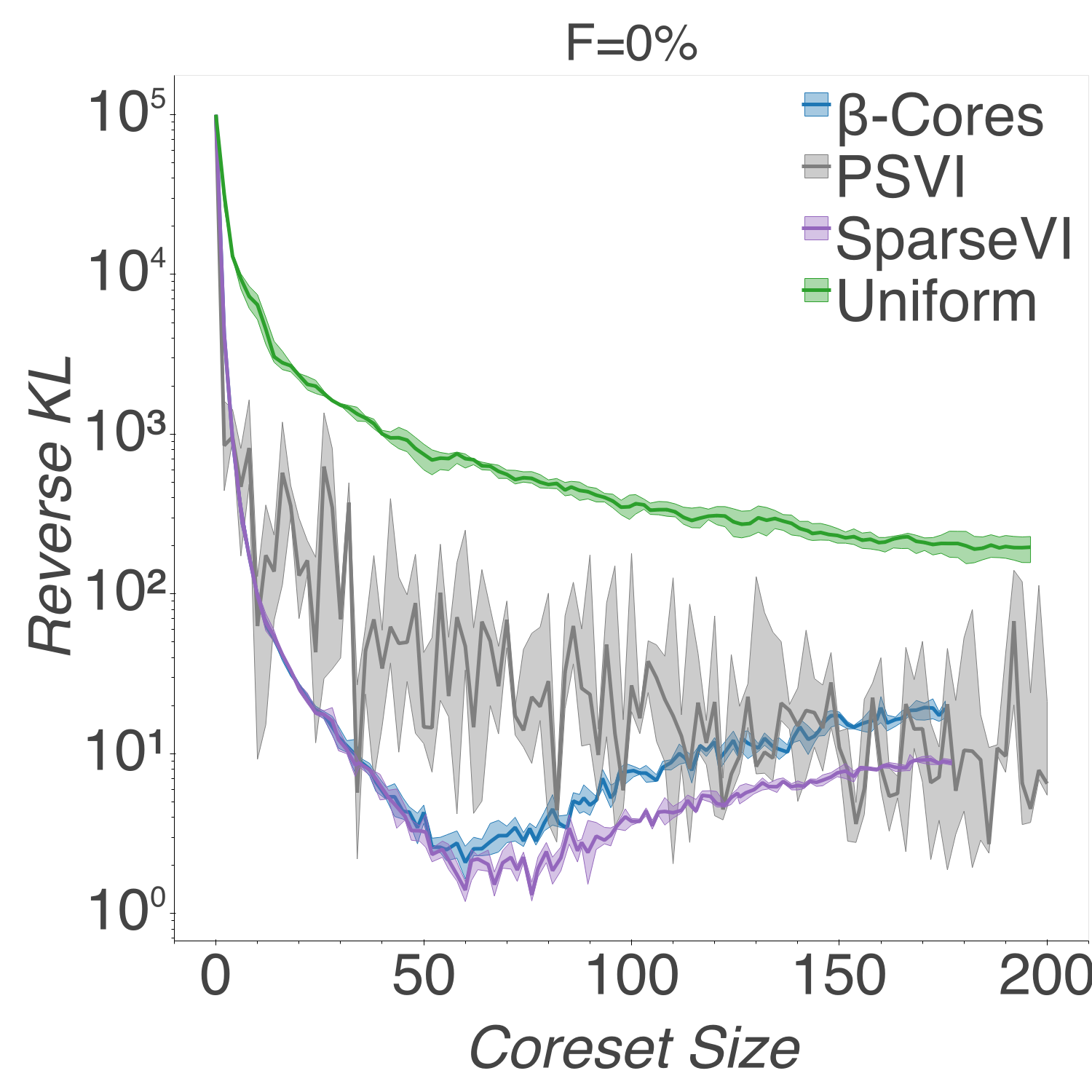}
		\centering
		\hfill
		\includegraphics[width=.3\textwidth]{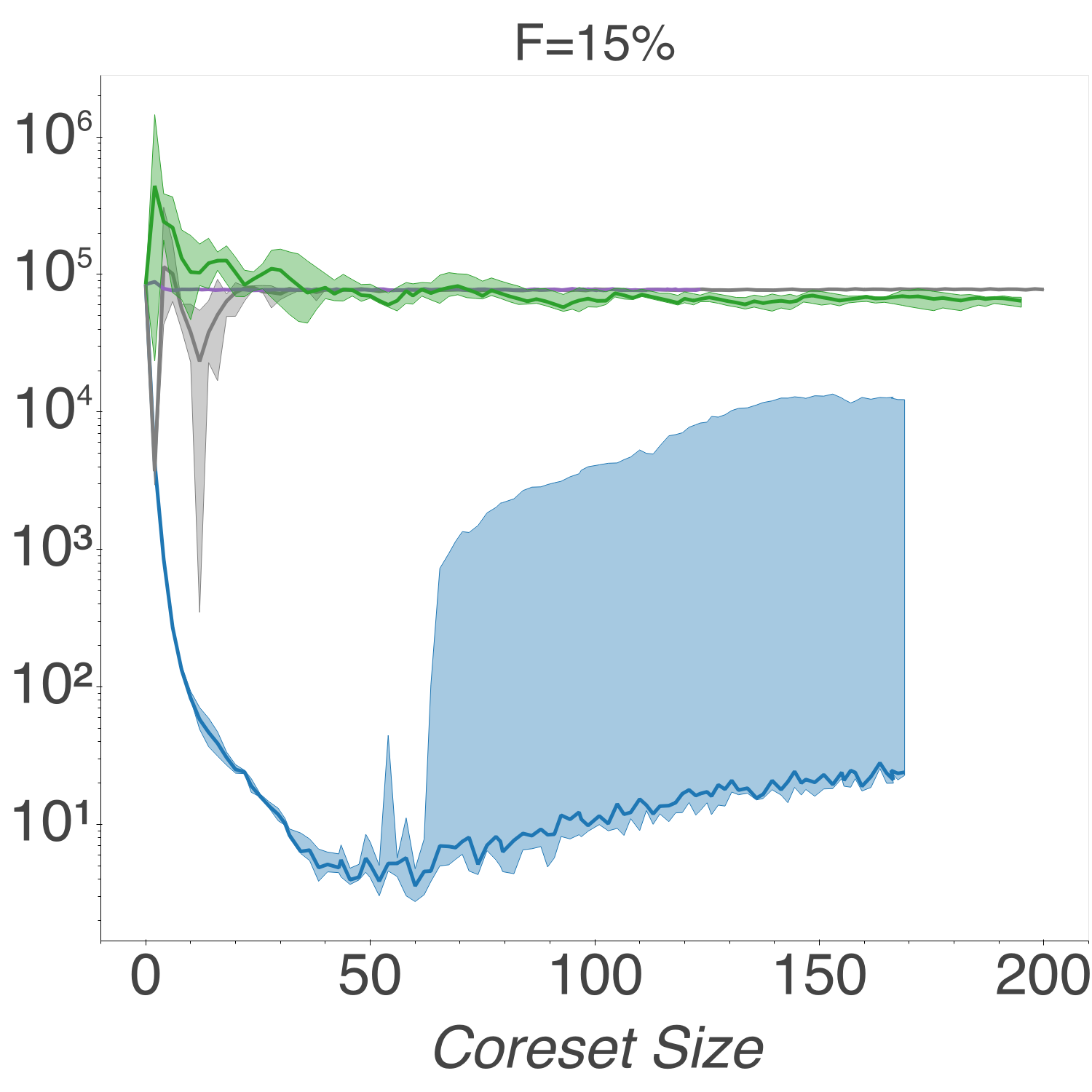}
		\centering
		\hfill
		\includegraphics[width=.3\textwidth]{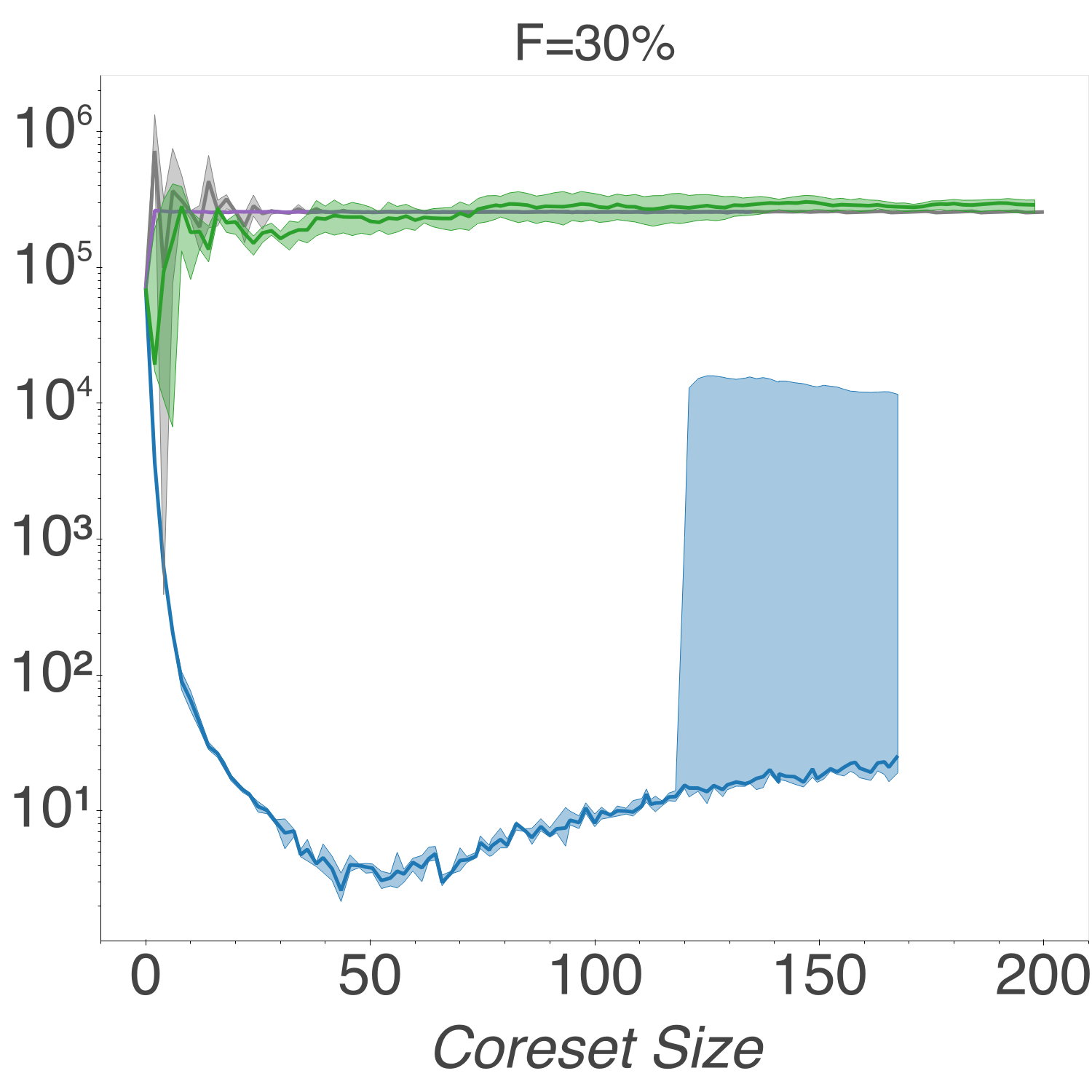}
		\caption{\label{fig:gauss_kld}}
	\end{subfigure}	
	\centering
	\caption{(a)~Scatterplot of the observed datapoints projected on two random axes, overlaid by the corresponding coreset points and predictive posterior $3\sigma$ ellipses for increasing coreset size (from left to right). Exact posterior (illustrated in black) is computed on the dataset after removing the group of outliers. From top to bottom, the level of structured contamination increases. Classical Riemannian coresets are prone to model misspecification, adding points from the outlying component, while \bcores{} adds points only from the uncontaminated subpopulation yielding better posterior estimation. (b) Reverse KL divergence between coreset and true posterior, averaged over $5$ trials. Solid lines display the median KL divergence, with shaded areas showing $25\textsuperscript{th}$ and $75\textsuperscript{th}$ percentiles of KL divergence.}
\end{figure}

All computations involved in the coreset construction and posterior evaluation in this experiment can be performed in closed form~\cite{campbell19neurips}. We apply the batch scheme of~\cref{alg:bcores}, sampling from the exact coreset posterior over gradient estimation. The used \mbox{($\beta$-)}likelihood equations are outlined in~\cref{sec:gauss-lik}. For all coreset methods, constructions are repeated for up to $M=200$ iterations, with $\gamma_t = t^{-1}$. Notice that our setting does not imply that maximum summary size contains 200 datapoints: often over the iterations an already existing summary point may be selected again, resulting in smaller coresets.

\cref{fig:gaussian_coreset_points}~presents the results obtained by the different coreset methods. We stress-test their performance under varying amounts of data corruption~(from top to bottom, 0\%, 15\%, and 30\% of the datapoints get replaced by outliers). We can verify that \bcores{} with $\beta=0.01$ is on par with existing Riemannian coresets in an uncontaminated dataset. Noticeably, \bcores{} remains robust to high levels of structured corruption~(even up to $30\%$ of the dataset), giving reliable posterior estimates; KL divergence plots in~\cref{fig:gauss_kld} reconfirm the superiority of inference via~\bcores{}. On the other hand, in the presence of outliers, previous Riemannian coresets performance degrades quickly, offering similar posterior inference quality with random sampling. The KL divergence from the cleansed data posterior for existing summarizations and uniform sampling increases with observations failure probability, as it asymptotically converges to the Bayesian posterior computed on the corrupted dataset. 

Moreover, in the case of contaminated datasets, baseline coresets are quite confident in their wrong predictive posteriors: they keep assigning the same weight to all observations and hence do not adjust their posterior uncertainty estimates, in spite of having to describe contradicting data. In contrast,~\bcores{} discards samples from the outlying group and can confidently explain the inliers, despite the smaller effective sample size: indeed,~\cref{fig:gauss_kld} shows that the achieved KL divergence from the exact posterior is at same order of magnitude regardless of failure probability. 

We can however notice that, for coreset sizes growing beyond 60 points---despite remaining consistently better compared to the baselines---\bcores{} starts to present some instability over trials in contaminated dataset instances. This effect is attributed to the small value of the $\beta$ hyperparameter  selected for the demonstration (so that this value can successfully model the case of clean data). As a result, eventually some outliers might be allowed to enter the summary for large coreset sizes. The instability can be resolved by increasing $\beta$ according to the observations failure probability.

\subsection{Bayesian Logistic Regression under Mislabeling and Feature Noise}
\label{subsec:logreg-expt}

In this section, we study the robustness achieved by~\bcores{} on the problem of binary classification  under unreliable measurements and labeling. We test our methods on 3 benchmark datasets with varying dimensionality~($10$-$127$ dimensions, more details on the data are provided in~\cref{sec:data-details}). We observe data pairs $(x_n, y_n)_{n=1}^{N}$, where $x\in\reals^{d}$, $y_n \in \{-1,1\}$, and use the Bayesian logistic regression model to describe them,
\[
y_n | x_n, \theta \sim \distBern \left( \frac{1}{1+e^{-z_n^T \theta}}\right),
\qquad 
z_n:=\begin{bmatrix}
x_n \\
1
\end{bmatrix}.
\label{eq:logreg-model}
\]
\blik{} terms required in our construction are computed in~\cref{sec:logreg-lik}. 

Data corruption is simulated by generating outliers in the input and output space similarly to~\cite{futami18}: For corruption rate $F$, we sample two random subsets of size $F\cdot N$ from the training data.  For the datapoints in the first subset, we replace the value of half of the features with Gaussian noise sampled \iid from $\distNorm(0,5)$; for the datapoints in the other subset, we flip the binary label. Over construction we use Laplace approximation~\cite{mackay03} to efficiently draw samples from the (non-conjugate) coreset posterior, while over evaluation coreset posterior samples are obtained via NUTS~\cite{hoffman14}. We evaluate accuracy over the test set, predicting labels according to the maximum log-likelihood rule under the posterior $\theta$ sampling distribution. Learning rate schedule was set to $\gamma_t=c_0 t^{-1}$, with $c_0$ set to 1 for \sparsevi{} and \bcores{}, and 0.1 for \psvi. 
The values for hyperparameter $\beta$ and learning rates $\gamma_t$ were chosen via cross-validation. 

\begin{figure}[t!]
	\begin{subfigure}[b]{0.9\textwidth} 
		\centering
		\includegraphics[width=.325\textwidth]{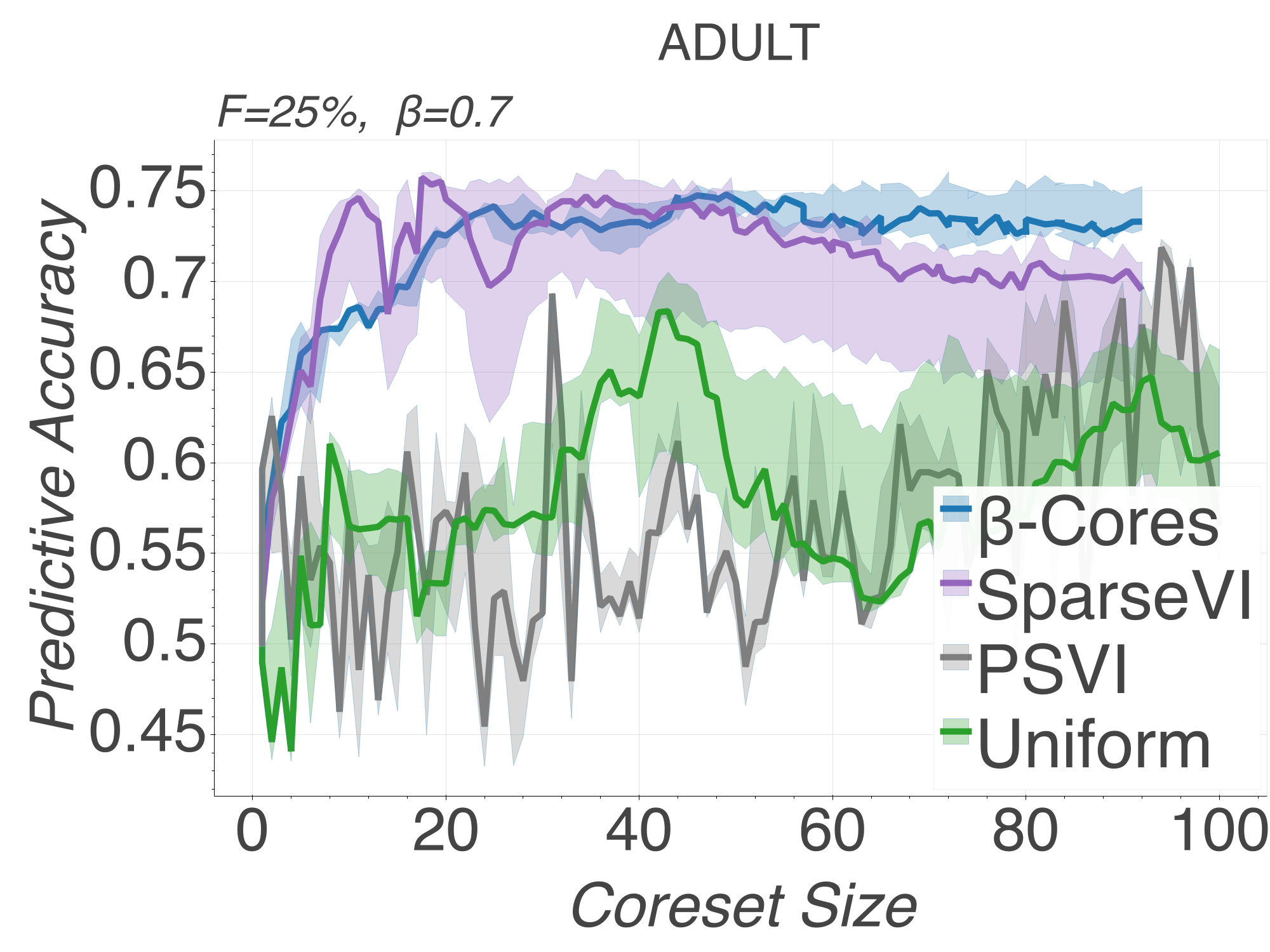}
		\centering
		\hfill
		\includegraphics[width=.325\textwidth]{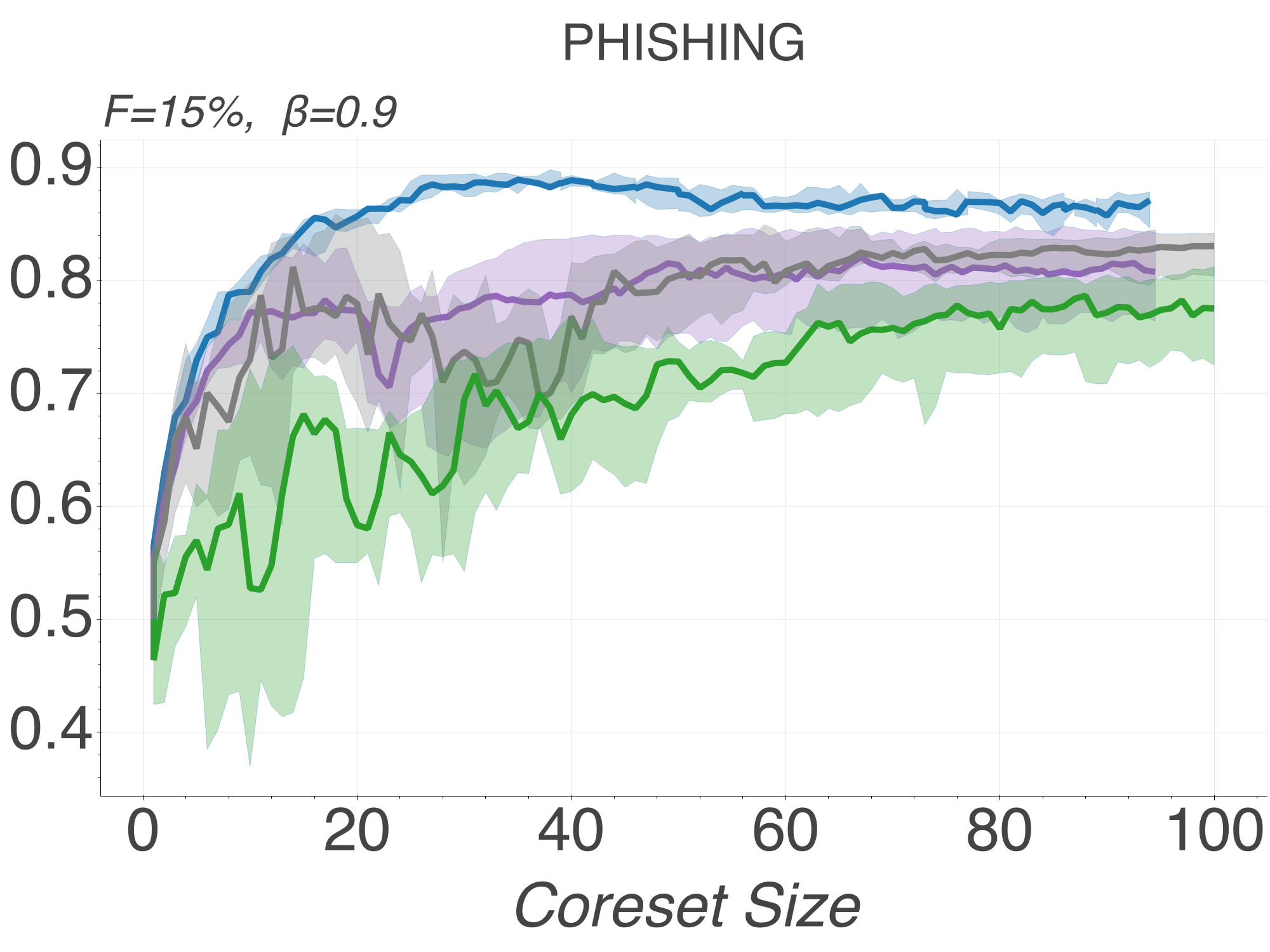}
		\centering
		\hfill
		\includegraphics[width=.325\textwidth]{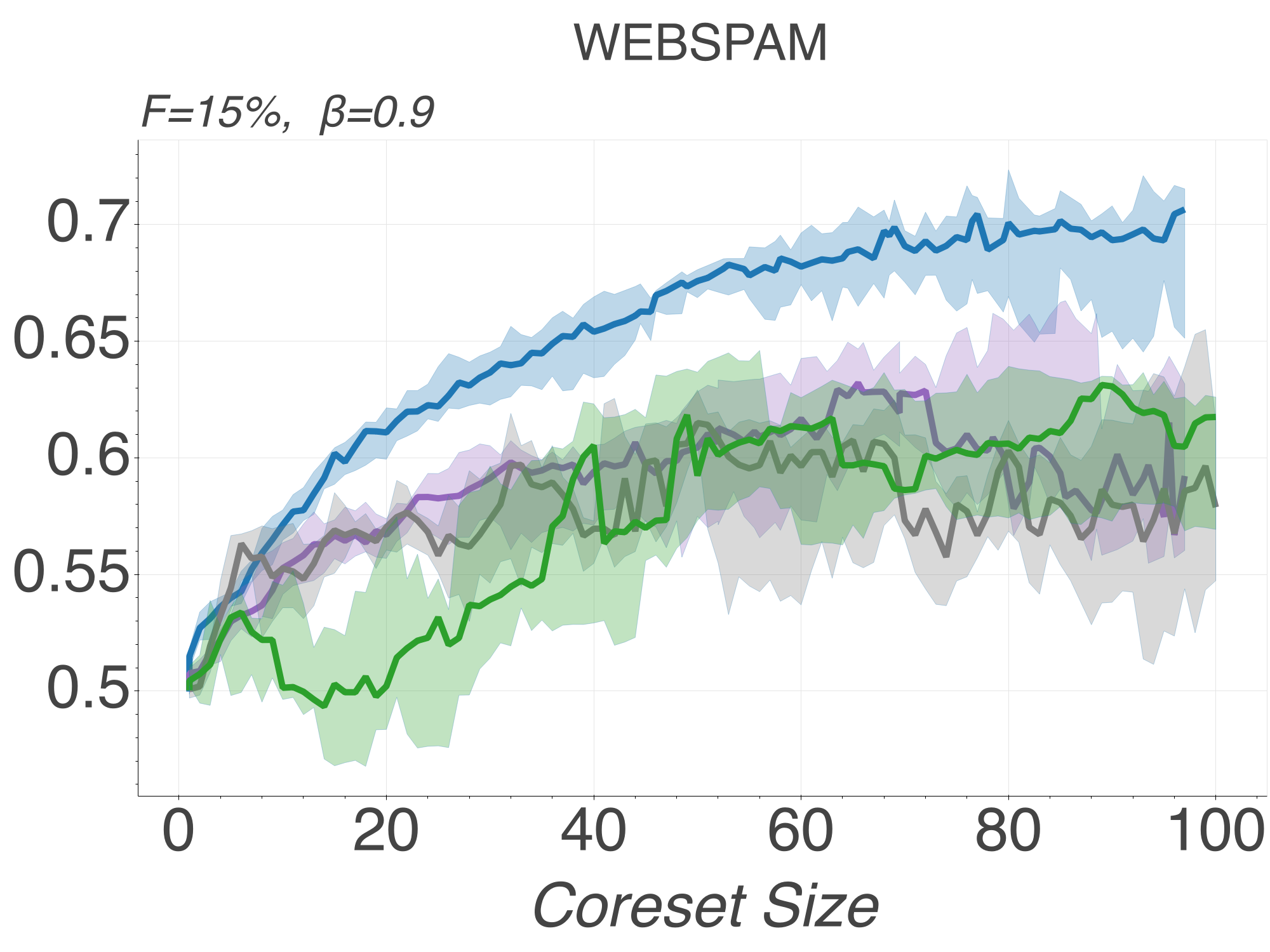}
		\centering
		\includegraphics[width=.325\textwidth]{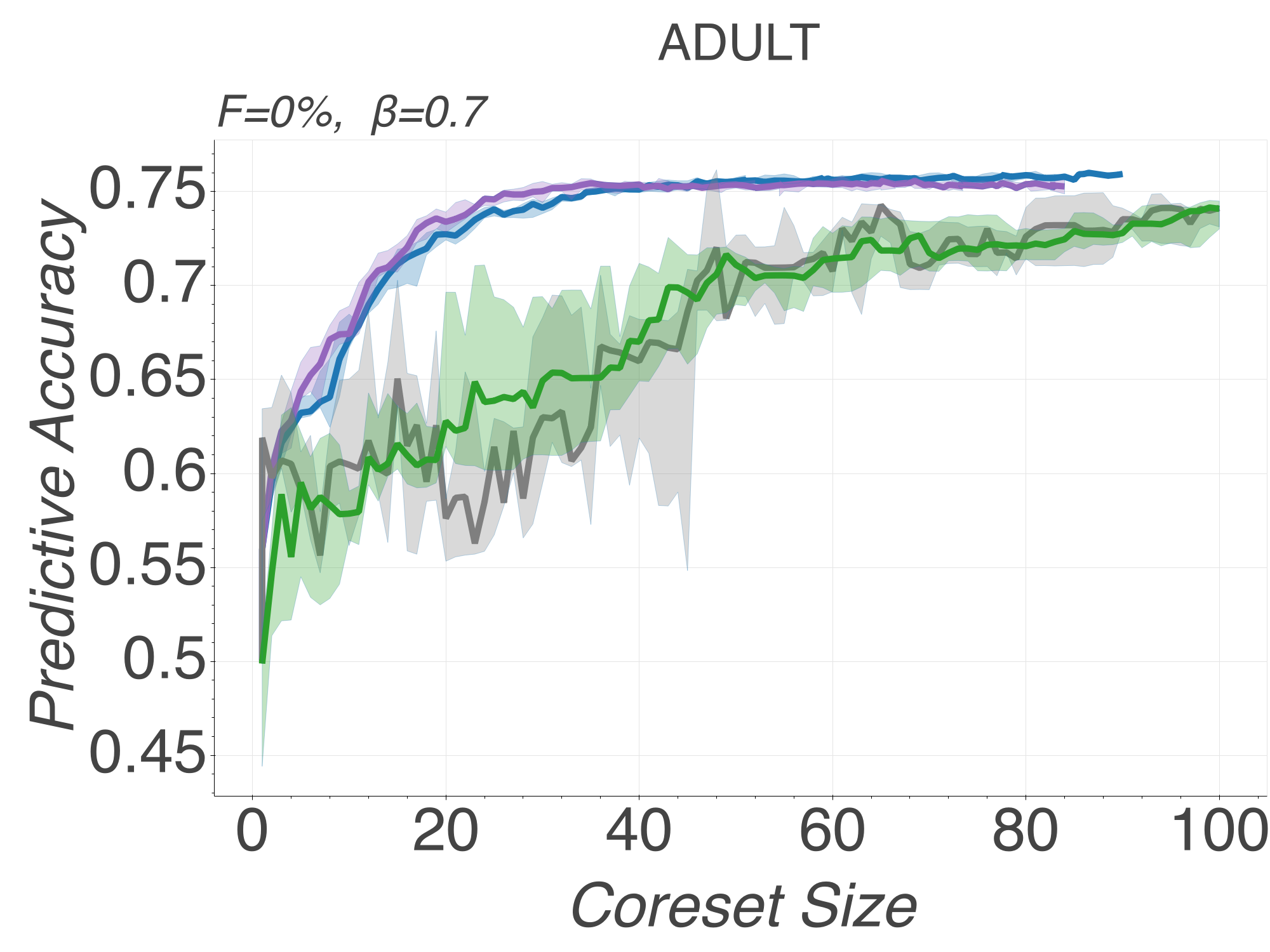}
		\centering
		\hfill
		\includegraphics[width=.325\textwidth]{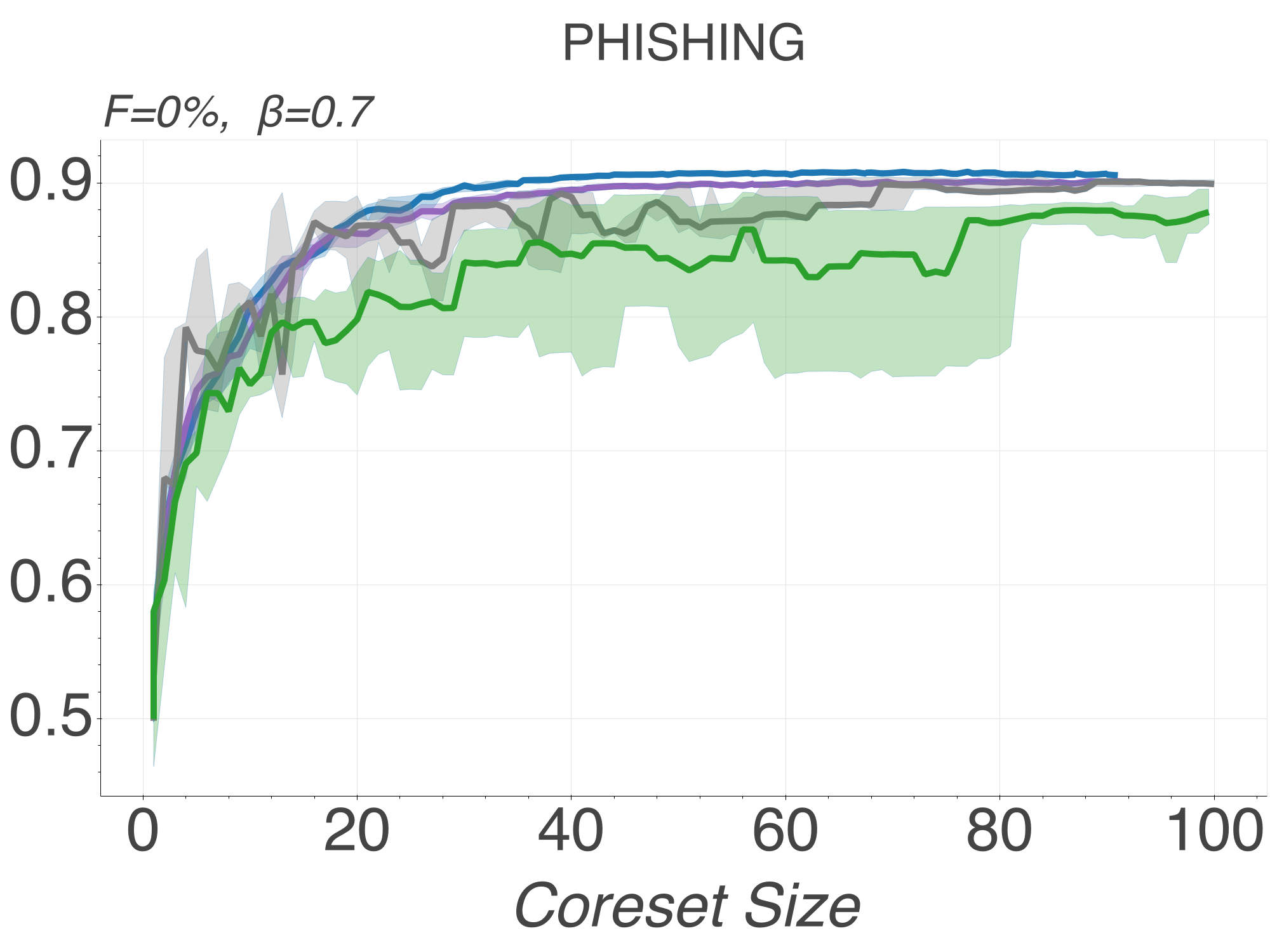}
		\centering
		\hfill
		\includegraphics[width=.325\textwidth]{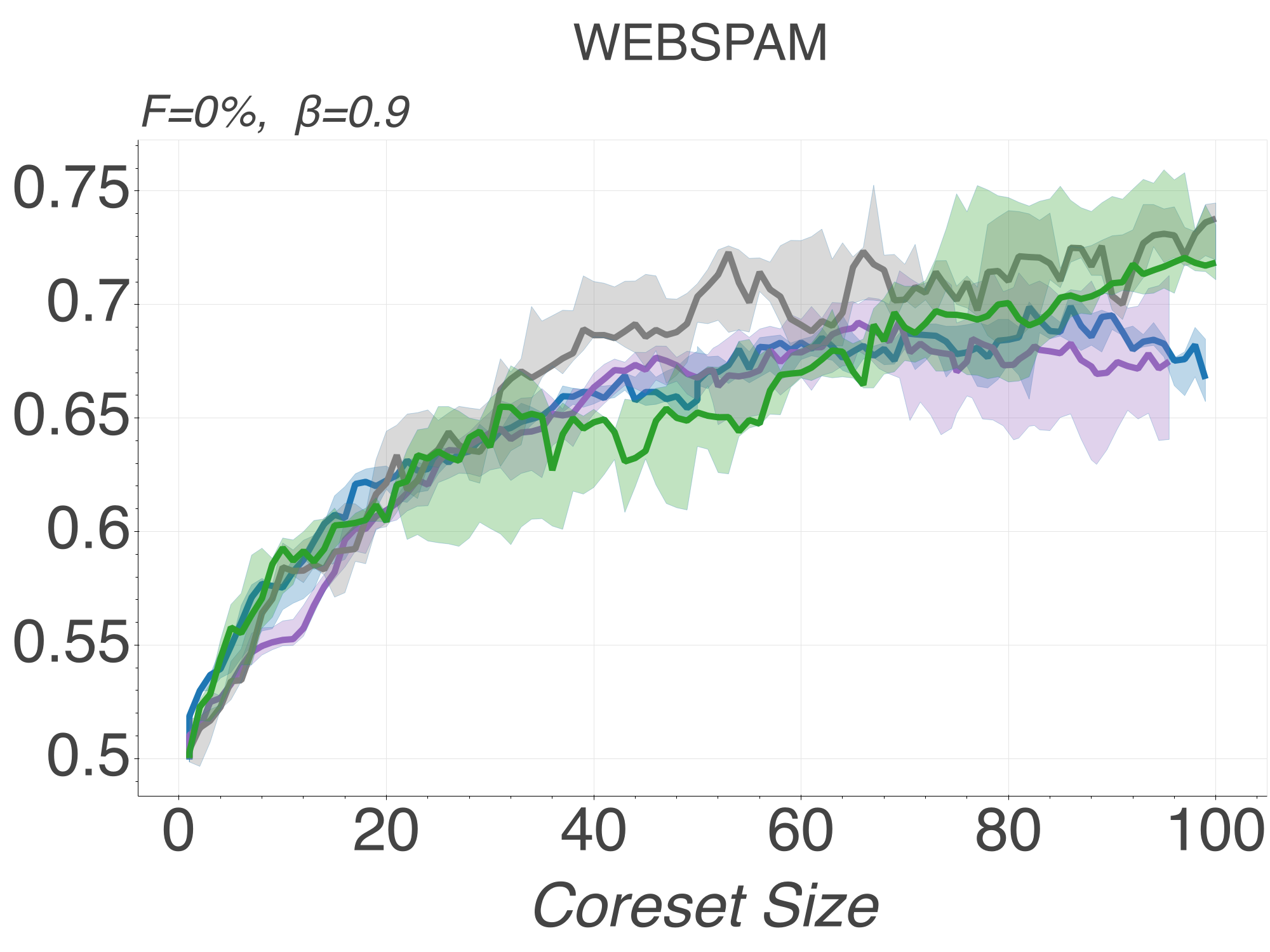}
	\end{subfigure}	
	\centering
	\caption{Predictive accuracy vs coreset size for logistic regression experiments over $10$ trials on $3$ large-scale datasets. Solid lines display the median accuracy, with shaded areas showing $25\textsuperscript{th}$ and $75\textsuperscript{th}$ percentiles. Dataset corruption rate $F$, and $\beta$ value used in \bcores{} for each experiment are shown on the figures. The bottom row plots illustrate the achieved predictive performance under no contamination.}
	\label{fig:logreg_plot}
\end{figure}

\cref{fig:logreg_plot} illustrates that \bcores{} shows competitive performance with the classic Riemannian coresets in the absence of data contamination~(bottom row), while it consistently achieves the best predictive accuracy in corrupted datasets~(top row).  On the other hand, ordinary summarization techniques, although overall outperforming random sampling for small coreset sizes, soon attain degraded predictive performance on poisoned data: by construction, via increasing coreset size, Riemannian coresets are expected to converge to the Bayesian posterior computed on the corrupted dataset. All baselines present noticeable degradation in their predictive accuracy when corruption is introduced (typically more than $5\%$), which is not the case for our method: \bcores{} is designed to support corrupted input and, for a well-tuned hyperparameter $\beta$, maintains similar performance in the presence of outliers, while practically it can even achieve improvement (as occurring for the \textsc{WebSpam} data).

\subsection{Neural Linear Regression on Noisy Data Batches}
\label{subsec:neur-linr-expt}

Here we use the coresets extension for batch summarization to efficiently train a neural linear model on selected data minibatches. Neural linear models perform Bayesian linear regression on the representation of the last layer of a deterministic neural network feature extractor~\cite{snoek15,riquelme18,pinsler19}.
The corresponding statistical model is as follows
\[
%&z(\cdot) \dist \distNorm(\mu_0, \sigma_0^2 I), \\
%\quad 
&\left(y_n\right)_{n=1}^{N} = \theta^T z(x_n) + \eps_n,
\quad
\left(\eps_n\right)_{n=1}^{N} \dist \distNorm(0, \sigma^2).
\label{eq:neurlinr-stat-model}
\]
The neural network is trained to learn an adaptive basis $z(\cdot)$ from $N$ datapoint pairs $(x_n,y_n) \in \reals^{d} \times \reals$, which we then use to regress $ \left(y_n\right)_{n=1}^{N} $ on $ \left(z(x_n)\right)_{n=1}^{N} $, and yield uncertainty aware estimates of $\theta$. More details on the model-specific formulae entering coresets construction are provided in~\cref{sec:neurlinr-lik}. Input and output related outliers are simulated as in~\cref{subsec:logreg-expt}, while here, for the output related outliers, $y_n$  gets replaced by Gaussian noise. Corruption occurs over a percentage $F\%$ of the total number of minibatches of the dataset, while the remaining minibatches are left uncontaminated. Each poisoned minibatch gets $70\%$ of its points \mbox{substituted by outliers}.

We evaluate \bcores, \sparsevi{} and random sampling on two benchmark regression datasets~(detailed in~\cref{sec:data-details}). All coresets are initialized to a small batch of datapoints sampled uniformly at random from the dataset inliers. Over incremental construction, we interleave each minibatch selection and weights optimization step of the coreset with a training round for the neural network, constrained on the current coreset datapoints. Each such training round consists of $10^3$ minibatch gradient descent steps using the AdaGrad optimizer~\cite{duchi11}.
Our neural architecture is comprised of two fully connected hidden layers, batch normalization and ReLU activation functions. The values of coreset size at initialization, batch size added per coreset iteration, and units at each neural network hidden layer are set respectively to 20, 10 and 30 for the~\textsc{Housing}, and 200, 100 and 100 for the~\textsc{Songs} dataset.

\begin{figure}[t!]
	\begin{subfigure}[b]{0.9\textwidth} 
		\centering
		\includegraphics[width=.45\textwidth]{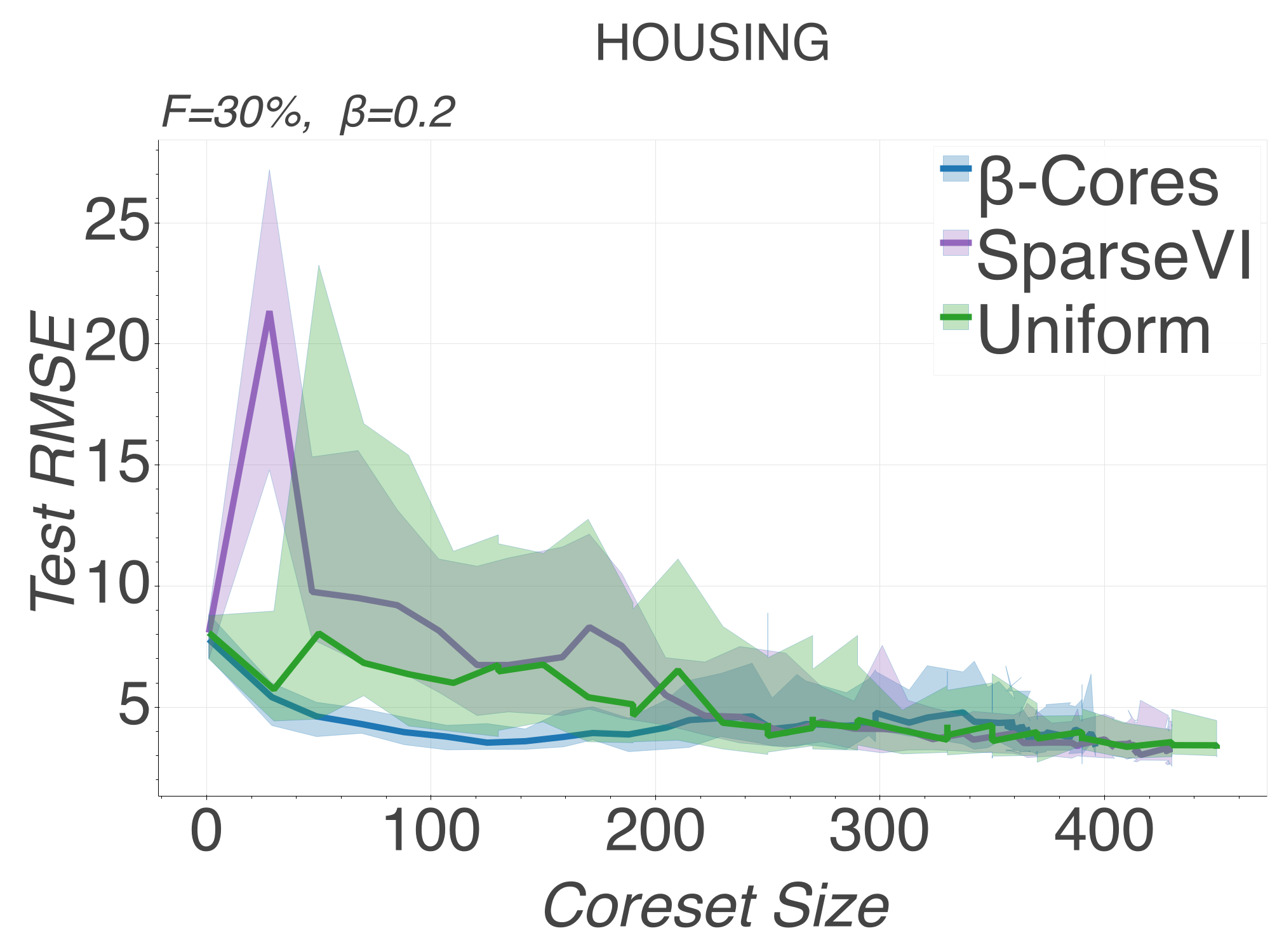}
		\hfill
		\includegraphics[width=.45\textwidth]{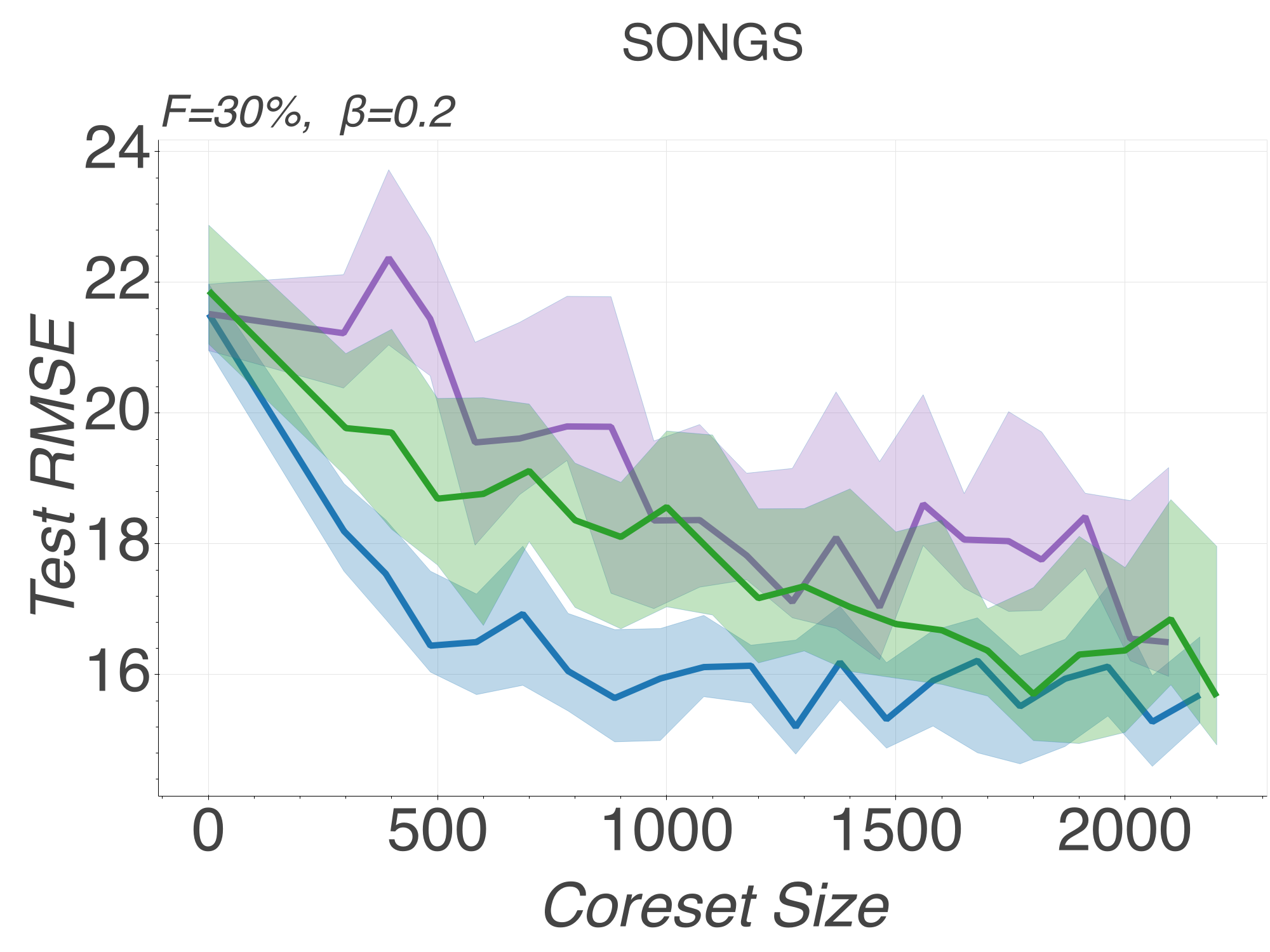}
		\centering
		\hfill
		\includegraphics[width=.45\textwidth]{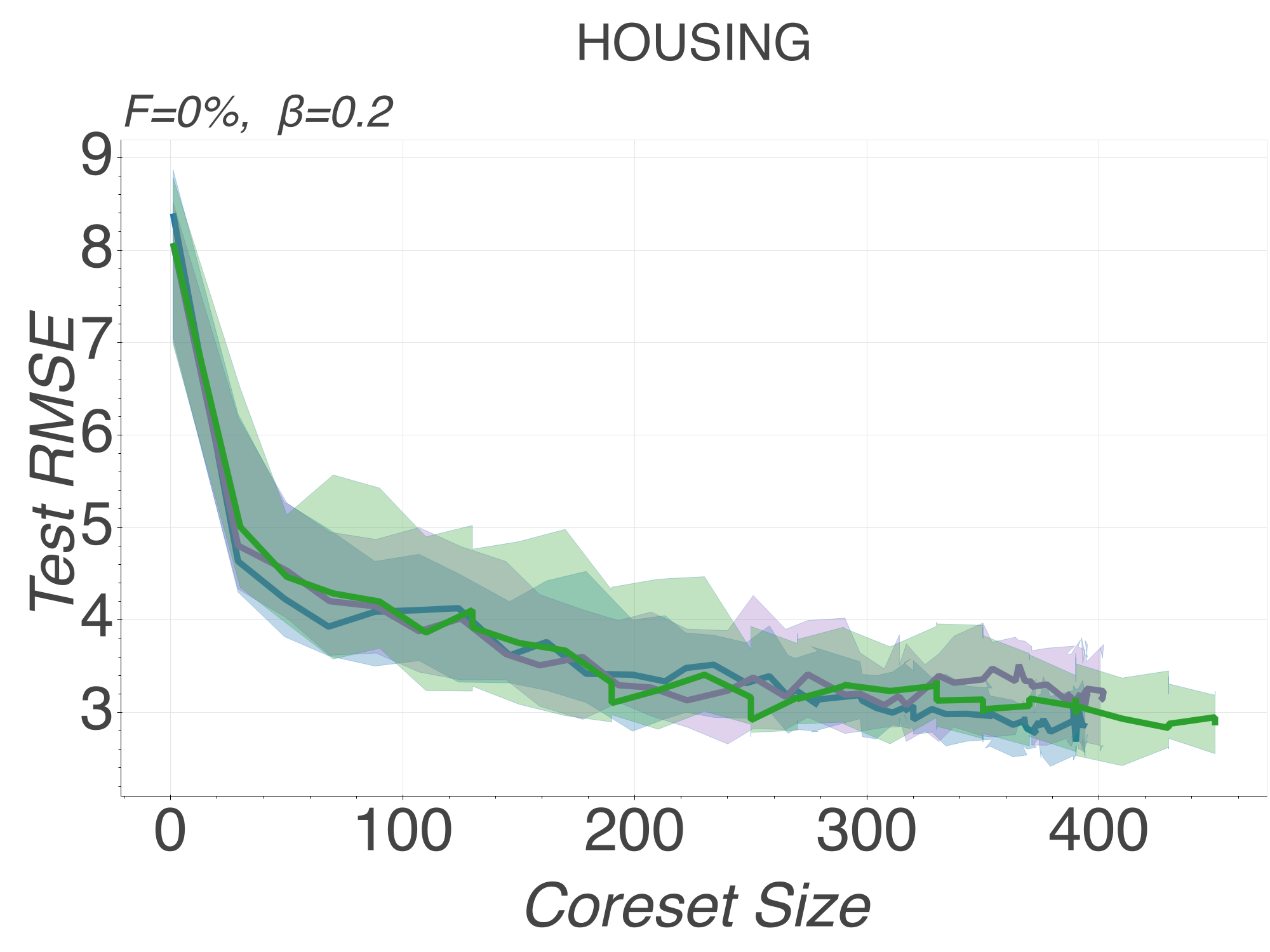}
		\centering
		\hfill
		\includegraphics[width=.45\textwidth]{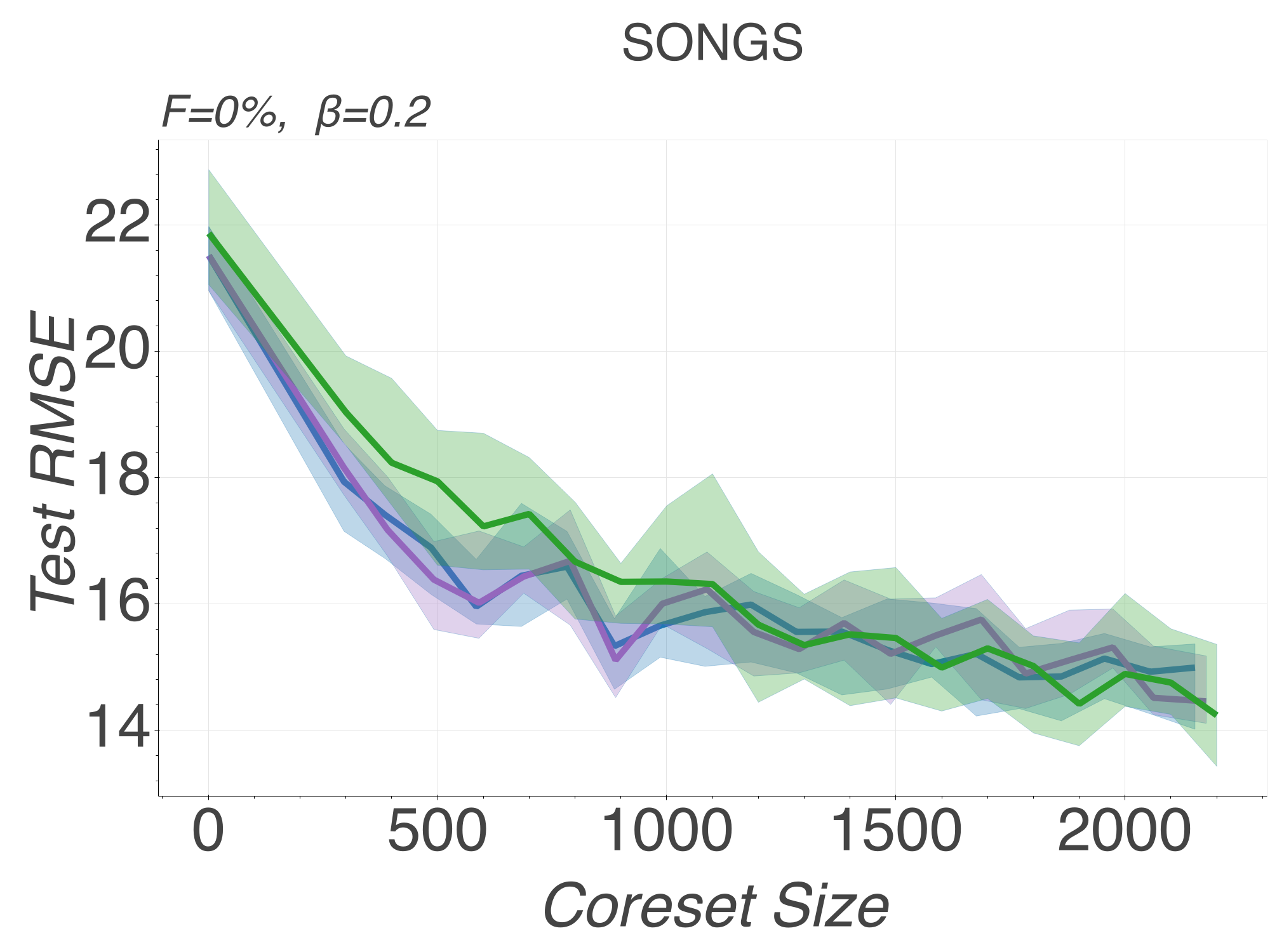}
	\end{subfigure}	
	\centering
	\caption{Test RMSE vs coreset size for neural linear regression experiments averaged over 30 trials. Solid lines display the median RMSE, with shaded areas showing $25\textsuperscript{th}$ and $75\textsuperscript{th}$ percentiles. Dataset corruption rate $F$, and $\beta$ value used in \bcores{} for each experiment are shown on the figures. The bottom row plots illustrate the achieved predictive performance under no contamination.}
	\label{fig:neural_plot}
\end{figure}

\cref{fig:neural_plot}~(bottom row) shows that \bcores{} are competitive with the baselines in the absence of data corruption, achieving similar predictive performance over the entire range of tested coreset sizes. Under data poisoning~(top row), \bcores{} is the only method that offers monotonic decrease of test RMSE for increasing summary size from the beginning of the experiment. On the other hand, baselines present unreliable predictive performance for small coreset sizes: random sampling and \sparsevi{} are both prone to including corrupted data batches, whose misguiding information gets expressed on the flexible representations learnt by the neural network, requiring a larger summary size to reach the RMSE of \bcores.

\subsection{Efficient Data Acquisition from Subpopulations for Budgeted Inference}
\label{sec:active-selection}

We consider the scenario where a machine learning service provider aims to fit a binary classification model to observations coming from multiple subpopulations of data contributors. The provider aims to maximize the predictive accuracy of the model, while adhering to a budget on the total number of subpopulations from which data can be used over inference. Budgeted inference can be motivated by several practical requirements: First, restricting the total number of datapoints used over learning to a smaller informative subset aids scalability---which is the primary motivation for coresets. Moreover, taking decisions at the subpopulations level regarding which groups of datapoints are useful for the task, without the need to inspect datapoints individually, reduces the privacy loss incurred over the data selection stage, and can be integrated in machine learning pipelines that follow formal hierarchical privacy schemes. Finally, subpopulations valuation can guide costly experimental procedures, via inducing knowledge regarding which group combinations are most beneficial in summarizing the entire population of interest~\citep{pinsler19, vahidian20}, and hence should be prioritised over data collection.

In this study we use a subset of more than $60K$ datapoints from the \textsc{HospitalReadmissions} dataset (for further details see \cref{sec:data-details}). Using combinations of age, race and gender information of data contributors, we form a total of $165$ subpopulations within the training dataset. Data contamination is simulated identically to the experiment of \cref{subsec:logreg-expt}, while now we also consider the case of varying levels of contamination across the subpopulations. In particular, we form groups of roughly equal size where $0\%, 10\%$ and $20\%$ of the datapoints get replaced by outliers---this results in getting a dataset with approximately $10\%$ of its full set of datapoints corresponding to outliers.

\begin{figure}[t!]
	\begin{subfigure}[b]{0.9\textwidth} 
		\centering
		\includegraphics[width=.45\textwidth]{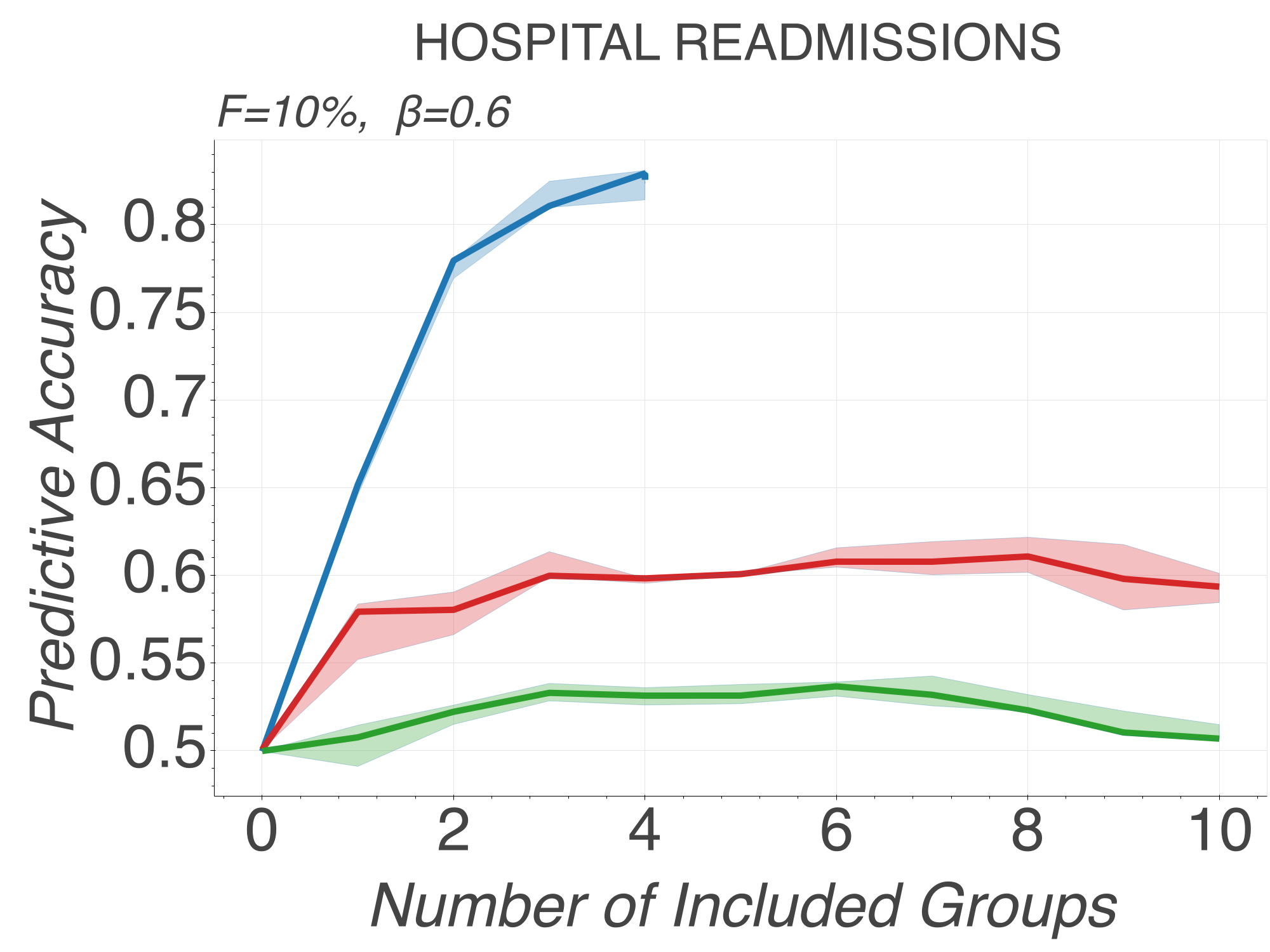}
		\hfill
		\includegraphics[width=.45\textwidth]{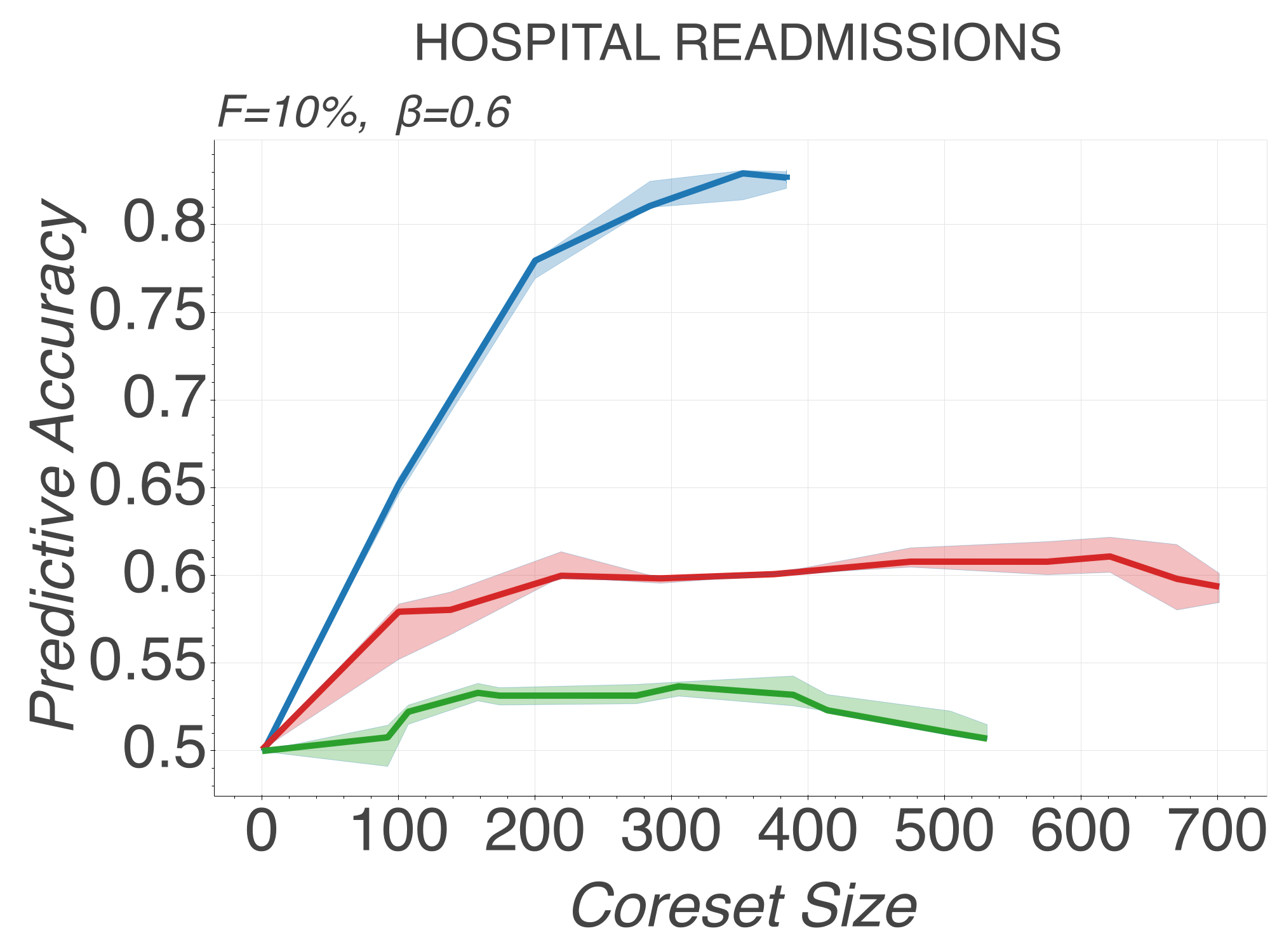}
		\centering
		\hfill
		\includegraphics[width=.45\textwidth]{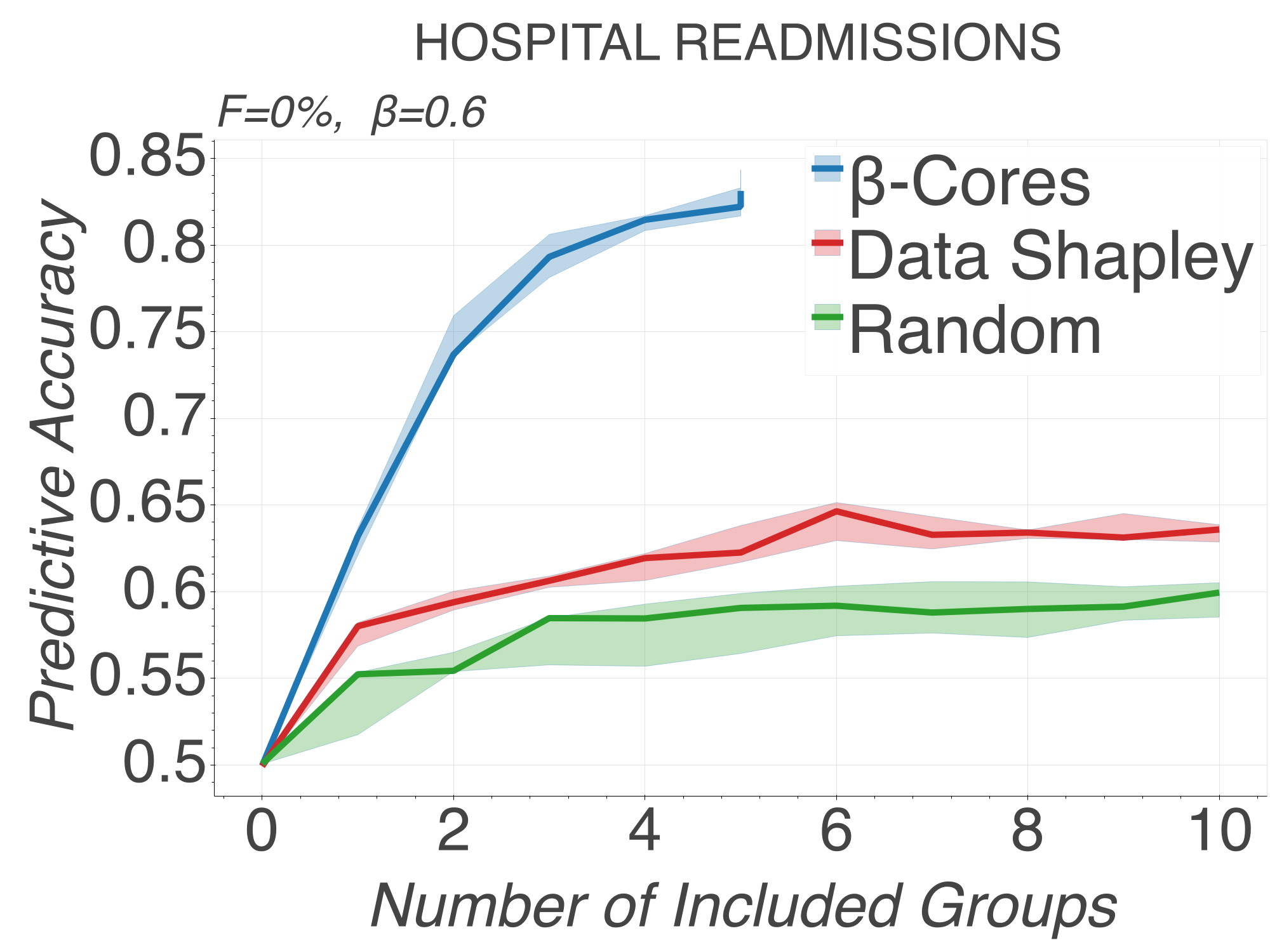}
		\centering
		\hfill
		\includegraphics[width=.45\textwidth]{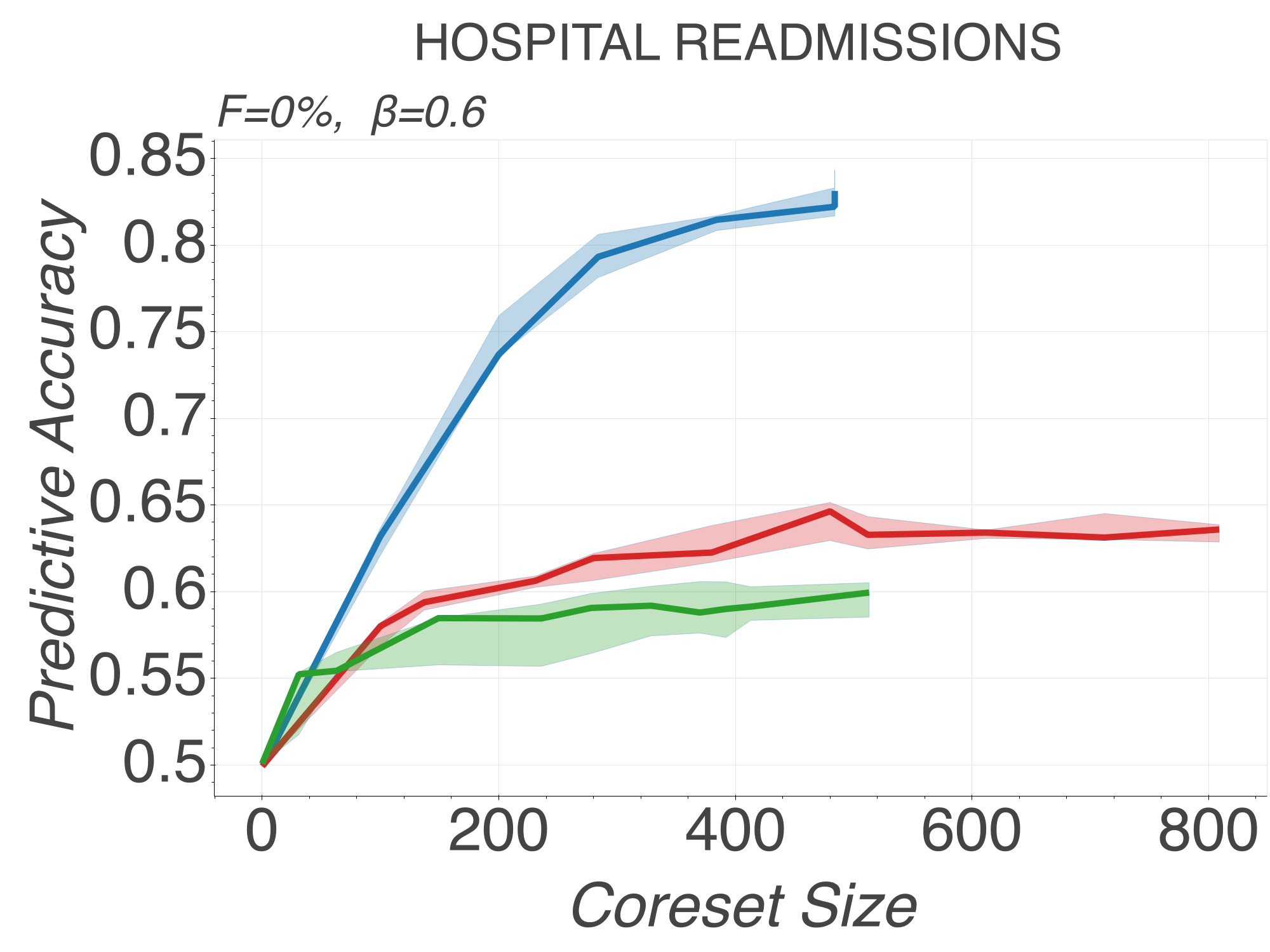}
	\end{subfigure}	
	\centering
	\caption{Predictive accuracy against number of groups~(left) and number of datapoints~(right) selected for inference. Compared group selection shemes are \bcores{}, selection according to Shapley values based ranking, and random selection. The experiment is repeated over $5$ trials, on a contaminated dataset containing a $10\%$ of crafted outliers distributed non-uniformly across groups~(top row), and a clean dataset~(bottom row).}
	\label{fig:group_plot}
\end{figure}

\begin{figure}[t!]
	\begin{subfigure}[b]{.8\textwidth} 
		\centering
		\includegraphics[width=.9\textwidth]{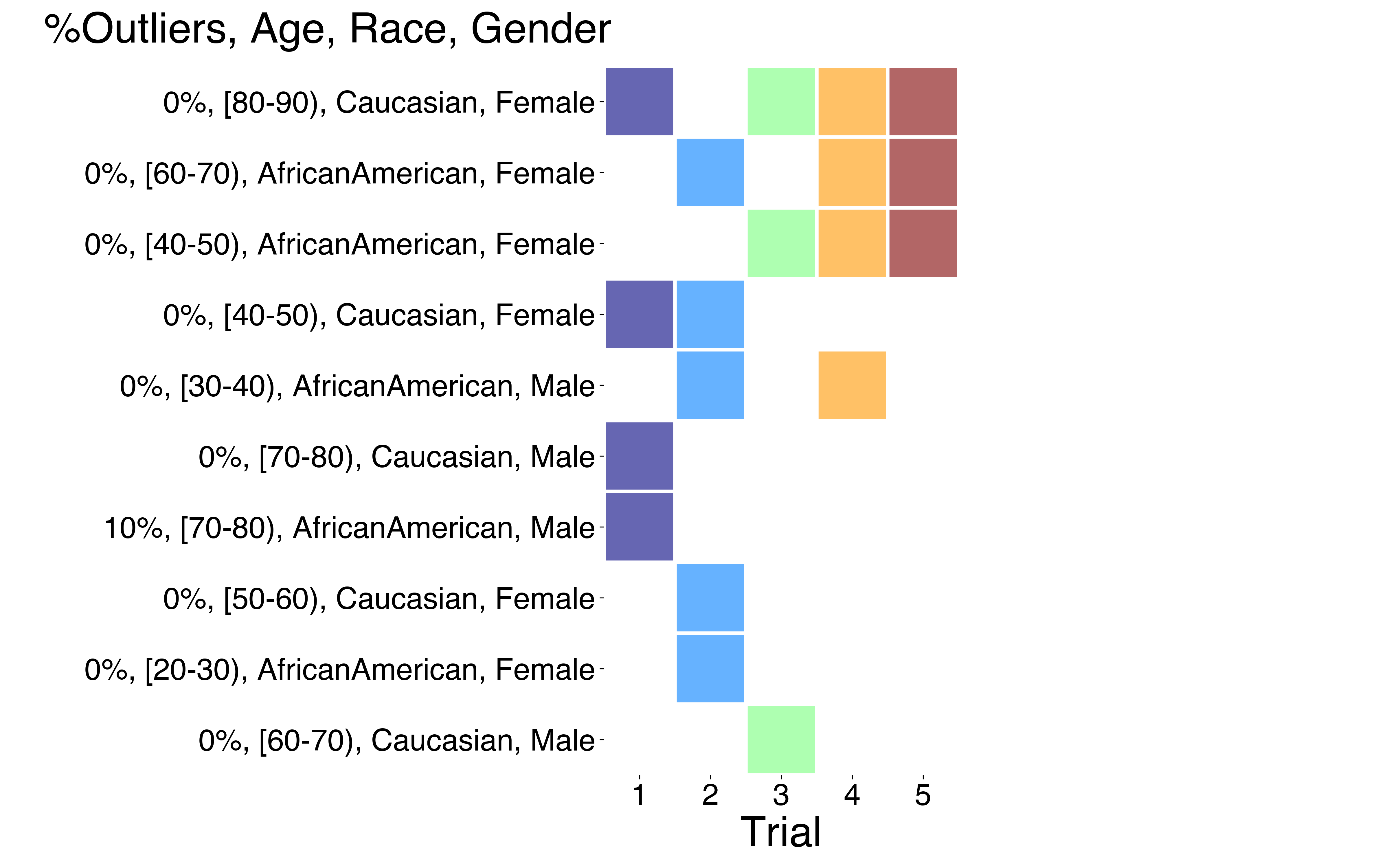}
	\end{subfigure}	
	\centering
	\caption{Attributes of selected groups after running $10$ iterations of \bcores{} with $\beta=0.6$ on the contaminated \textsc{HospitalReadmissions} dataset (repeated over $5$ random trials).}
	\label{fig:selected_groups}
\end{figure}

We evaluate the predictive accuracy achieved by doing inference on the data subset obtained after running $10$ iterations of the \bcores{} extension for groups (which gives a maximum of $10$ selected groups). We compare against (\emph{i}) a random sampler, and (\emph{ii}) a baseline which ranks all groups according to their Shapley value and selects the groups with the highest values. Shapley value is a concept originating in cooperative game theory~\citep{shapley53}, which has recently found applications in data valuation and outliers detection~\cite{ghorbani19}. In the context of our experiment, it quantifies what is the marginal contribution of each group to the predictive accuracy of the model at all possible group coalitions that can be formed. As this quantity is notoriously expensive to be computed in large datasets, we use a Monte Carlo estimator which samples $5K$ possible permutations of groups and for each permutation it computes marginals for coalitions formed by the first $20$ groups.\footnote{The latter truncation is supported by the observation that marginal contributions to the predictive accuracy are diminishing as the dataset size increases.}

As illustrated in~\cref{fig:group_plot}, \bcores{} with $\beta=0.6$ offers the best solution to our problem, and is able to reach predictive accuracy exceeding $75\%$ by fitting a coreset on no more than $2$ groups. ~\cref{fig:selected_groups} displays the demographic information of selected groups. We can notice that subpopulations of female and older patients are more informative for the classification task, while Caucasian and African-American groups are preferred to smaller racial minorities. Importantly, \bcores{} is able to distill clean from contaminated groups. For used $\beta$ value we can see than over the set of trials only one group with outliers level of $10\%$ is allowed to enter a summary, which already contains $3$ uncontaminated groups.

Shapley values based ranking treats outliers better than random sampling: As outliers are expected to have negative marginal contribution to predictive accuracy, their Shapley rank is generally lower compared to clean data groups. On the other hand, Shapley computation is much slower \mbox{than random} sampling and \bcores, specific to the evaluation metric of interest, while Shapley values are not designed to find data-efficient combinations of groups, hence this baseline can still return redundancy in the selected \mbox{data subset}.

\section{Conclusion \& further directions}
\label{sec:conclusion}
In this work, we proposed a general purpose framework for yielding  contami\-nation-robust summarizations of massive scale datasets for inference. Relying on recent advances in Bayesian coresets and robustified inference under the \bdiv{}, we developed a greedy black-box construction that efficiently shrinks big data via keeping informative datapoints, while simultaneously rejecting outliers.
 Finally, we presented experiments involving various statistical models, and simulated and real-world datasets, demonstrating that our methodology outperforms existing techniques in scenarios of structured and unstructured data corruption. 

Our future work will be concerned with considering stronger adversarial settings where summaries are initialized to data subsets that already contain outliers. Further directions also include automating the tuning of the robustness hyperparameter $\beta$, as well as applying our techniques to more complicated statistical models, including ones with structured likelihood functions (e.g. time-series and temporal point processes).
\section{Acknowledgements}
\label{section:acknowledgements}

This work is partially supported by Nokia Bell Labs through their donation for the Centre of Mobile, Wearable Systems and Augmented Intelligence to the University of Cambridge. DM also gratefully acknowledges the support received from Lundgren Fund and Darwin College Cambridge. We thank Trevor Campbell for helpful discussions on Bayesian coresets.

\appendix
\section{Models}
\label{sec:models}
In this section we present the derivations of \blik{} terms~\cref{eq:b-loss,eq:sl-lik-terms} required over the \bcores{} constructions for the statistical models of our experiments.

\subsection{Gaussian likelihoods}
\label{sec:gauss-lik}

For the \blik{} terms of a multivariate normal distribution, we have 
\[
\pi(x|\mu, \Sigma)^{\beta} = \left((2\pi)^{-\frac{d}{2}}|\Sigma|^{-\frac{1}{2}}\right)^{\beta} \exp\left(-\frac{\beta}{2}(x-\mu)^T\Sigma^{-1}(x-\mu)\right),
\]
and, by simple calculus (see also~\citep{samek13}),
\[
\int_{\mcX}\pi(\chi|\mu, \Sigma)^{1+\beta}d\chi = \left((2\pi)^{-\frac{d}{2}}|\Sigma|^{-\frac{1}{2}}\right)^{\beta}(1+\beta)^{-\frac{d}{2}}.
\]
Hence
\[
f_n(\mu) 
 \propto &  \frac{1}{\beta}\left((2\pi)^{-\frac{d}{2}}|\Sigma|^{-\frac{1}{2}}\right)^{\beta} \exp\left(-\frac{\beta}{2}(x-\mu)^T\Sigma^{-1}(x-\mu)\right) \\
 &-\left((2\pi)^{-\frac{d}{2}}|\Sigma|^{-\frac{1}{2}}\right)^{\beta}(1+\beta)^{-\frac{d}{2}-1}\\
 \propto &
 \frac{1}{\beta} \exp\left(-\frac{\beta}{2}(x-\mu)^T\Sigma^{-1}(x-\mu)\right) 
 -(1+\beta)^{-\frac{d}{2}-1}
 \label{eq:gaussian-beta-lik}.
\]

\subsection{Logistic regression likelihoods}
\label{sec:logreg-lik}
Log-likelihood terms of individual datapoints are given as follows
\[
\log \pi(y_n|x_n, \theta) = -\log\left(1+e^{-y_n z_n^T \theta}\right).
\label{eq:logreg-loglik}
\]
Substituting to~\cref{eq:sl-lik-terms}, for the  \blik{} terms we get
\[
f_n(\theta)& \propto -\frac{1}{\beta}\left(1+e^{-y_n z_n^T \theta}\right)^{-\beta} \\
&+ \frac{1}{\beta+1} \left( \left(1+e^{- z_n^T \theta}\right)^{-(\beta+1)} + \left(1+e^{z_n^T \theta}\right)^{-(\beta+1)} \right).
\label{eq:logreg-blik}
\]

\subsection{Neural linear regression likelihoods and predictive posterior}
\label{sec:neurlinr-lik}
Recall that in the neural linear regression model, $ \left(y_n - \theta^T z(x_n)\right) \sim \distNorm(0, \sigma^2)$, $n=1,\ldots,N$.
Then the Gaussian log-likelihoods corresponding to individual observations (after dropping normalization constants),  are written as 
\[
f_n(\theta) = - \frac{1}{2\sigma^2}\left(y_n - \theta^T z(x_n)\right)^2.
\label{eq:neurlinr-logliks}
\]
Assuming a prior $\theta \dist \distNorm(\mu_0, \sigma_0^2 I)$, the coreset posterior can be computed in closed form as follows
\[
\pi_w(\theta) = \distNorm\left(\mu_w, \Sigma_w\right),
\label{eq:neurlinr-coreset-posterior}
\]
where 
\[
&\Sigma_w := \left(\sigma_0^{-2}I + \sigma^{-2} \sum_{m=1}^{M}w_m z(x_m) z(x_m)^T \right)^{-1},\\
&\mu_w := \Sigma_w \left( \sigma_0^{-2} I \mu_0 + \sigma^{-2} \sum_{m=1}^{M} w_m y_m z(x_m) \right).
\label{eq:neurlinr-corest-posterior-params}
\]
By substitution to~\cref{eq:sl-lik-terms},
the \blik{} terms for our adaptive basis linear regression are written as 
\[
f_n(\theta) \propto  \frac{1}{(2\pi)^{\beta/2}\sigma^{\beta}} \left(-\frac{\beta+1}{\beta}e^{-\beta\left(y_n-\theta^Tz(x_n)\right)^2/(2\sigma^2)} + \frac{1}{\sqrt{1+\beta}}\right).
\label{eq:linreg-blik}
\]
%To simplify notation 
Let $\mcC$ be the output of the coreset applied on a dataset $\mcD$. Hence, in regression problems, the predictive posterior on a test data pair $(x_t, y_t)$ via a coreset is approximated as follows
\[
\pi(y_t|x_t, \mcD) & \approx \pi(y_t|x_t, \mcC)  \\
&= \int \pi(y_t|x_t,  \theta) \pi(\theta|\mcC) d\theta.  
\label{eq:coreset-postpred}
\]
In the neural linear experiment, 
the predictive posterior is a Gaussian given by the following formula
\[
\pi(y_t|x_t, \mcC) 
& = \distNorm \left(y_t; \mu_w^T z(x_t), \sigma^2 + z(x_t)^T \Sigma_w z(x_t)\right).
\label{eq:neurlinr-pred-posterior}
\]

\section{Datasets Details}
\label{sec:data-details}

The benchmark datasets used in logistic regression (including group selection) and neural linear regression experiments are detailed in Tables~\ref{table:logreg-data} and \ref{table:neurlinreg-data} respectively.\footnote{The original versions of all used datasets can be accessed by following the corresponding hyperlinks in the Tables appearing in the electronic version of the paper.}, and include: 
\begin{itemize}
\item a dataset used to predict whether a citizen's income exceeds $50K \$$ per year extracted from USA 1994 census data~(\textsc{Adult}),
\item a dataset containing webpages features and a label categorizing them as phishing or not~(\textsc{Phishing}),
\item a corpus of webpages crawled from links found in spam emails~(\textsc{WebSpam}),
\item a set of hospitalization records for binary prediction of readmission pertaining to diabetes patients~(\textsc{HospitalReadmissions}),
\item a set of various features from homes in the suburbs of Boston, Massachussets used to model housing price~(\textsc{Housing}), and
\item a dataset used to predict the release year of songs from associated audio features ~(\textsc{Songs}).
\end{itemize} 

For \textsc{Adult}, \textsc{Phishing} and \textsc{HospitalReadmissions} we fit our statistical models on the first 10 principal components of the datasets, while  all logistic regression benchmark datasets are evaluated on balanced subsets of the test data between the two classes~(see~\cref{table:logreg-data}).
	\begin{table}[!t]
		\caption{Logistic regression datasets}
		\centering
		\resizebox{\columnwidth}{!}{%
		\begin{tabular}{lrrrrr}
			\hline
			Dataset      &   $d$ &     $N\textsubscript{train}$ &   $N\textsubscript{test}$ &   $\#$Pos. test data 
			 \\
			\hline
			\MYhref{http://archive.ics.uci.edu/ml/datasets/Adult}{\textsc{Adult}}~\citep{adult}        &  10 &  30,162 &    7,413 &             3,700  \\
			\MYhref{https://archive.ics.uci.edu/ml/datasets/Phishing+Websites}{\textsc{Phishing}}~\citep{uci}      & 10 & 8,844 &   2,210 &             1,230  \\
			 \MYhref{https://www.cc.gatech.edu/projects/doi/WebbSpamCorpus.html}{\textsc{WebSpam}}~\citep{webspam}      & 127 & 126,185 &   13,789 &             6,907  \\
			\MYhref{https://archive.ics.uci.edu/ml/datasets/diabetes+130-us+hospitals+for+years+1999-2008}{\textsc{HospitalReadmissions}}~\citep{diabetes}      & 10 & 55,163 &   6,079 &             3,044  \\
			\hline
		\end{tabular}
	}
		\label{table:logreg-data}
\end{table}

	\begin{table}[!t]
		\caption{Neural linear regression datasets}
		\centering
		\resizebox{0.6\columnwidth}{!}{%
		\begin{tabular}{lrrrrr}
			\hline
			Dataset      &   $d$ &      $N\textsubscript{train}$  &   $N\textsubscript{test}$   \\
			\hline
			\MYhref{https://archive.ics.uci.edu/ml/machine-learning-databases/housing/}{\textsc{Housing}}~\citep{uci}  &  13 &  446 &    50  \\
			\MYhref{https://archive.ics.uci.edu/ml/datasets/YearPredictionMSD}{\textsc{Songs}}~\citep{uci}        &  90 &  463,711 &    51,534\\
			\hline
		\end{tabular}
	}
		\label{table:neurlinreg-data}
	\end{table}

\clearpage
%\bibliography{ms}

\end{document}